\documentclass[sigconf]{acmart}

\usepackage{subfigure}
\usepackage{multirow}
\usepackage{xspace}
\usepackage{listings}
\usepackage{xcolor}
\usepackage{makecell}
\usepackage{CJKutf8}

\newcommand{\sys}{\textsc{GAMLP}\xspace}

\newcommand{\blue}[1]{\textcolor{blue}{#1}}

\AtBeginDocument{%
  \providecommand\BibTeX{{%
    \normalfont B\kern-0.5em{\scshape i\kern-0.25em b}\kern-0.8em\TeX}}}

\copyrightyear{2022}
\acmYear{2022}
\setcopyright{acmcopyright}\acmConference[KDD '22]{Proceedings of the 28th ACM SIGKDD Conference on Knowledge Discovery and Data Mining}{August 14--18, 2022}{Washington, DC, USA}
\acmBooktitle{Proceedings of the 28th ACM SIGKDD Conference on Knowledge Discovery and Data Mining (KDD '22), August 14--18, 2022, Washington, DC, USA}
\acmPrice{15.00}
\acmDOI{10.1145/3534678.3539121}
\acmISBN{978-1-4503-9385-0/22/08}

\begin{document}
\fancyhead{}
\title{Graph Attention Multi-Layer Perceptron}

\author{Wentao Zhang$^{1,3}$, Ziqi Yin$^{4}$, Zeang Sheng$^{1}$, Yang Li$^{1}$, Wen Ouyang$^{3}$, \\Xiaosen Li$^{3}$, Yangyu Tao$^{3}$, 
Zhi Yang$^{1,2}$, 
Bin Cui$^{1,2}$}
\affiliation{
% $^\dagger$Peking University\country{China}~~~~~
 {{$^1$}{School of CS \& Key Laboratory of High Confidence Software Technologies, Peking University}}\\
 {{$^{2}$}{Center for Data Science, Peking University \& National Engineering Laboratory for Big Data Analysis and Applications}}\\
$^3$Tencent Inc.\country{}~~~~~
 {{$^{4}$}{Beijing Institute of Technology}}
}
\affiliation{
$^1$\{wentao.zhang,  shengzeang18, liyang.cs, shenyu, yangzhi, bin.cui\}@pku.edu.cn\\ $^3$\{wtaozhang, hansenli, gdpouyang, brucetao\}@tencent.com~~~~~$^4$\{ziqiyin18\}@bit.edu.cn\country{}
}
\renewcommand{\shortauthors}{WT Zhang, et al.}
\renewcommand{\authors}{Wentao Zhang, Ziqi Yin, Zeang Sheng, Yang Li, Wen Ouyang, Xiaosen Li, Yangyu Tao, Zhi Yang, and Bin Cui}
\renewcommand{\shortauthors}{Wentao Zhang, et al.}

\begin{abstract}
Graph neural networks (GNNs) have achieved great success in many graph-based applications. However, the enormous size and high sparsity level of graphs hinder their applications under industrial scenarios.
Although some scalable GNNs are proposed for large-scale graphs, they adopt a fixed $K$-hop neighborhood for each node, thus facing the over-smoothing issue when adopting large propagation depths for nodes within sparse regions.
To tackle the above issue, we propose a new GNN architecture --- Graph Attention Multi-Layer Perceptron (GAMLP), which can capture the underlying correlations between different scales of graph knowledge.
We have deployed \sys in Tencent with the Angel platform~\footnote{\blue{\url{https://github.com/Angel-ML/PyTorch-On-Angel}}}, and we further evaluate GAMLP on both real-world datasets and large-scale industrial datasets.
Extensive experiments on these 14 graph datasets demonstrate that GAMLP achieves state-of-the-art performance while enjoying high scalability and efficiency. 
Specifically, it outperforms GAT by 1.3\% regarding predictive accuracy on our large-scale Tencent Video dataset while achieving up to $50\times$ training speedup.
Besides, it ranks top-1 on both the leaderboards of the largest homogeneous and heterogeneous graph (i.e., ogbn-papers100M and ogbn-mag) of Open Graph Benchmark~\footnote{\blue{\url{{https://ogb.stanford.edu/docs/leader\_nodeprop}}}}.
\end{abstract}

% <concept_id>10002950.10003624.10003633.10010917</concept_id>
% <concept_desc>Mathematics of computing~Graph algorithms</concept_desc>
% <concept_id>10010147.10010257.10010293.10010294</concept_id>
% <concept_desc>Computing methodologies~Neural networks</concept_desc>
% \ccsdesc[500]{Mathematics of computing~Graph algorithms}
% \ccsdesc[500]{Computing methodologies~Neural networks}

\begin{CCSXML}
<ccs2012>
<concept>
<concept_id>10002950.10003624.10003633.10010917</concept_id>
<concept_desc>Mathematics of computing~Graph algorithms</concept_desc>
<concept_significance>500</concept_significance>
</concept>
</ccs2012>
\end{CCSXML}

\ccsdesc[500]{Mathematics of computing~Graph algorithms}
\keywords{Scalable Graph Neural Network, Attention}


\maketitle

\section{Introduction}
Graph Neural Networks (GNNs) have been widely used in many tasks, including node classification, link prediction, and recommendation~\citep{zhang2020reliable, zhang2021rod, cui2020adaptive, miao2021degnn,jiang2022zoomer}.
Many industrial graphs are sparse, thus requiring GNNs to leverage long-range dependencies to enhance the node embeddings from distant neighbors.
Through stacking $K$ layers, GNNs can learn node representations by utilizing information from $K$-hop neighborhoods~\cite{miao2021lasagne}. 
The node set composed of the nodes within the $K$-hop neighborhood of a specific node is called this node's Receptive Field (RF).
However, as RF grows exponentially with the number of GNN layers, the rapidly expanding RF incurs high computation and memory costs in a single machine. 
Besides, even in the distributed environment, GNNs still have to pull features from a massive number of neighbors to compute the embedding of each node, leading to high communication cost~\citep{distdgl_ai3_2020}.
Due to the high computation and communication cost, most GNNs are hard to scale to large-scale industrial graphs.

A commonly used method to tackle the scalability issue (i.e., the recursive neighborhood expansion) in GNNs is sampling, and the sampling strategies have been widely researched~\cite{hamilton2017inductive, DBLP:conf/iclr/ChenMX18} and applied in many industrial GNN systems~\cite{distdgl_ai3_2020, aligraph_vldb_2019}. However, the sampling-based methods are imperfect because they still face high communication costs, and the sampling quality highly influences the model performance. As a result, many recent advancements towards scalable GNNs are based on model simplification~\citep{wu2019simplifying,zhang2021node,zhang2022pasca}, orthogonal to the sampling-based methods.

For example, Simplified GCN (SGC)~\citep{wu2019simplifying} decouples the feature propagation and transformation operation, and the former one is executed during pre-processing.
Unlike the sampling-based methods, which execute feature propagation during each training epoch, this time-consuming process in SGC is only executed once, and only the nodes of the training set are involved in the training process.
Therefore, SGC is computation and memory-efficient in a single machine and scalable in distributed settings.
Despite the high efficiency and scalability, SGC only preserves a fixed RF for all the nodes by assigning them the same feature propagation depth.
This fixed propagation mechanism in SGC disables its ability to exploit knowledge within neighborhoods of different sizes.

\begin{figure*}[tp!]
\vspace{-5mm}
\centering  
\subfigure[Inconsistent optimal steps.]{
\label{fig:ob1}
\scalebox{0.6}{
   \includegraphics[width=1\linewidth]{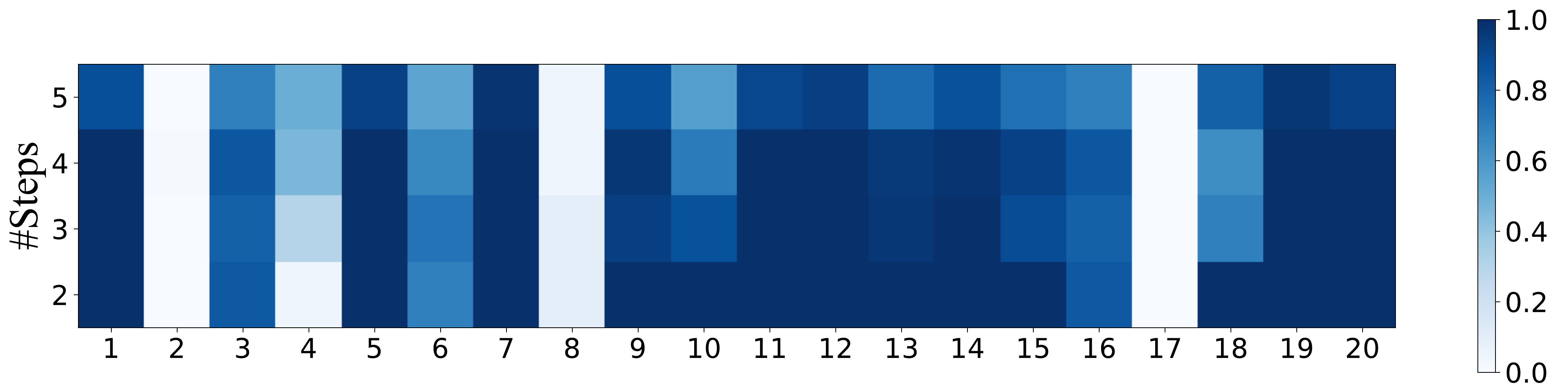}
 }}
\centering
\subfigure[Inconsistent RF expansion speed.]{
\label{fig:ob2}
\scalebox{0.3}{
   \includegraphics[width=1\linewidth]{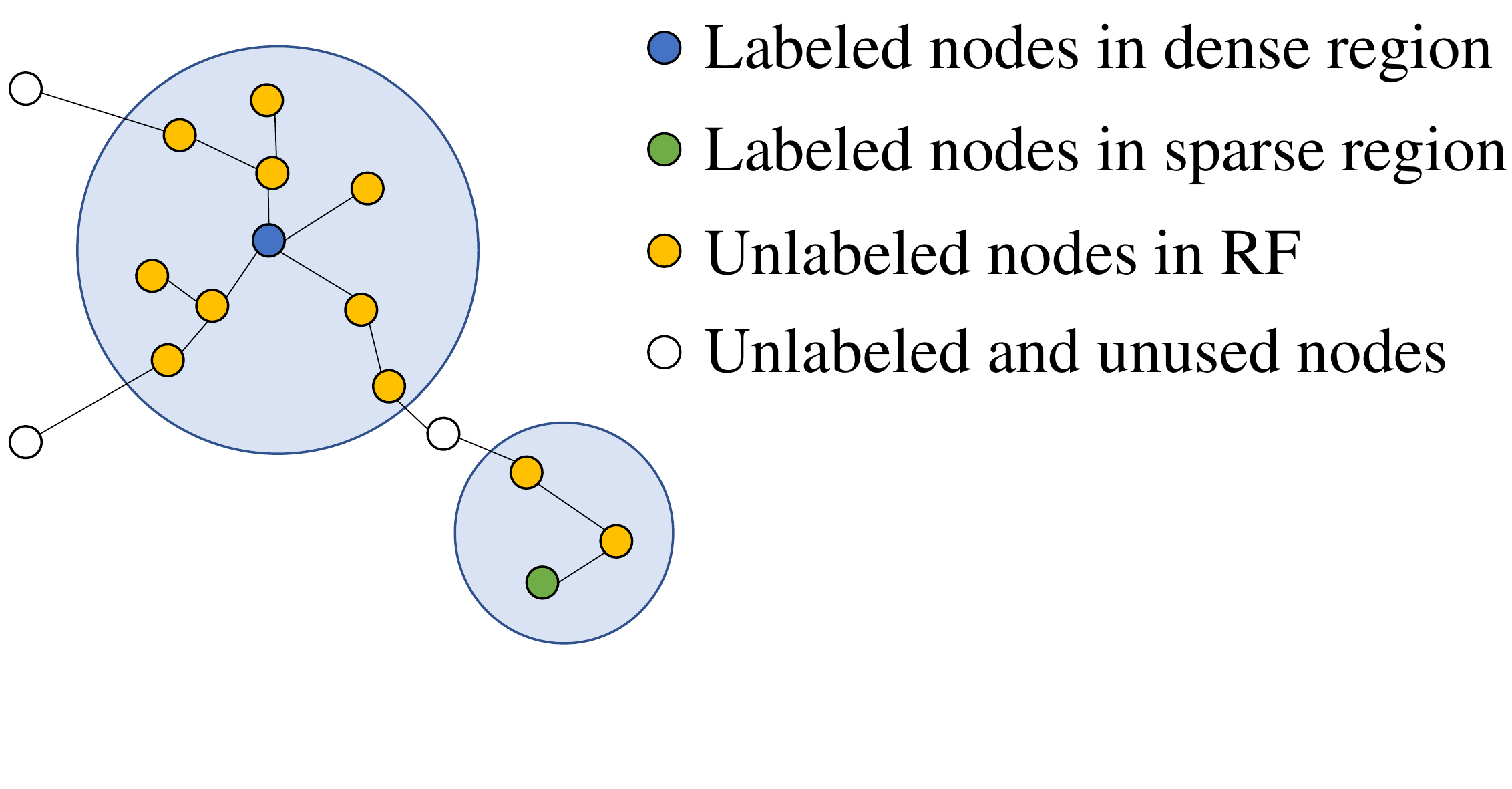}
 }}
\vspace{-4mm}
\caption{  (Left) Test accuracy of SGC on 20 randomly sampled nodes of Citeseer. The X-axis is the node id, and Y-axis is the propagation steps. The color from white to blue represents the ratio of being predicted correctly in 50 different runs. (Right) The node in the dense region has a larger RF within two iterations of propagation.
}
\label{fig.observation}
\vspace{-4mm}
\end{figure*}

Lines of simplified GNNs are proposed to tackle the fixed RF issue in SGC.
SIGN~\citep{frasca2020sign} concatenates all the propagated features without information loss, while S$^2$GC~\citep{zhu2021simple} averages all these propagated features to generate the combined feature. 
Although multi-scale knowledge is considered, the importance and correlations between different scales are ignored in these methods. 
To fill this gap, GBP~\citep{DBLP:conf/nips/ChenWDL00W20} adopts a heuristic constant decay factor for the weighted average for propagated features at different propagation steps.
Motivated by Personalized PageRank, the features with a larger propagation step has a higher risk of over-smoothing~\citep{li2018deeper,xu2018representation}, and they will contribute less to the combination in GBP. 

Unfortunately, the coarse-grained, layer-wise combination prevents these methods from unleashing their full potential.
As shown in Fig.~\ref{fig:ob1}, different nodes require different propagation steps to achieve optimal predictive accuracy. 
Besides, assigning the same weight distribution to propagated features along with propagation depth to all the nodes may be unsuitable due to the inconsistent RF expansion speed shown in Fig.~\ref{fig:ob2}.
However, nodes in most existing GNNs are restricted to a fixed-hop neighborhood and insensitive to the actual demands of different nodes. 

Motivated by the above observations, we propose to explicitly learn the importance and correlation of multi-scale knowledge in a node-adaptive manner. 
To this end, we develop a new architecture -- Graph Attention Multi-Layer Perceptron (GAMLP) -- that could automatically exploit the knowledge over different neighborhoods at the granularity of nodes.
GAMLP achieves this by introducing two novel attention mechanisms: \emph{Recursive attention} and \emph{Jumping Knowledge (JK) attention}. 
These two attention mechanisms can capture the complex correlations between propagated information at different propagation depths in a node-adaptive manner.
Consequently, \sys has the same benefits as the existing simplified and scalable GNN models while providing much better performance derived from its ability to utilize a node-adaptive RF.
Moreover, the proposed attention mechanisms can be applied to both node features and labels over neighborhoods with different sizes. 
By combining these two categories of information, GAMLP could achieve the best of both worlds in terms of accuracy.

Our contributions are as follows: 
(1) \textit{\underline{New perspective.}} We propose GAMLP, a scalable, efficient, and deep graph model. To the best of our knowledge, GAMLP is the first to explore both node-adaptive feature and label propagation schemes in scalable GNNs.
% (2) \textit{\underline{Novel method.}} 
(2) \textit{\underline{Real world deployment and applications.}} We deploy GAMLP in a distributed training manner in Tencent, and it has been widely used to support many applications in the real-world production environment.
(3) \textit{\underline{State-of-the-art performance.}} Experimental results demonstrate that GAMLP achieves state-of-the-art performance on 14 graph datasets while maintaining high scalability and efficiency.
For example, \sys outperforms GAT by 1.3\% regarding predictive accuracy on our large-scale Tencent Video dataset while achieving up to $50\times$ training speedup.
Besides, it outperforms the competitive baseline GraphSAINT~\citep{DBLP:conf/iclr/ZengZSKP20} in terms of accuracy by a margin of $0.42\%$, $3.02\%$ and $0.44\%$ on PPI, Flickr, and Reddit datasets under the inductive setting.
Under the transductive setting in large OGB datasets, the accuracy of \sys exceeds the second-best method by $1.03\%$, $1.32\%$ $1.61\%$ on the ogbn-products, ogbn-papers100M, and ogbn-mag datasets, respectively. 

\section{Preliminaries}
\subsection{Problem Formulation} 
We consider an undirected graph $\mathcal{G}$ = ($\mathcal{V}$,$\mathcal{E}$) with $|\mathcal{V}| = n$ nodes, $|\mathcal{E}| = m$ edges, and $c$ different node classes. We denote by $\mathbf{A}$ the adjacency matrix of $\mathcal{G}$, weighted or not. 
Each node has a feature vector of size $f$, stacked up in an $n \times f$ matrix $\mathbf{X}$. $\mathbf{D}=\operatorname{diag}\left(d_{1}, d_{2}, \cdots, d_{n}\right) \in \mathbb{R}^{n \times n}$ denotes the degree matrix of $\mathbf{A}$, where $d_{i}=\sum_{v_{j} \in \mathcal{V}} \mathbf{A}_{i j}$ is the degree of node $v_{i}$. 
Suppose $\mathcal{V}_l$ is the labeled set, and our goal is to predict the labels for nodes in the unlabeled set $\mathcal{V}_u$ with the supervision of $\mathcal{V}_l$.

\subsection{Scalable GNNs}
\textbf{Sampling.} 
As a node-wise sampling method, GraphSAGE~\citep{hamilton2017inductive} randomly samples a fixed-size set of neighbors for computation in each mini-batch. VR-GCN~\citep{DBLP:conf/icml/ChenZS18} analyzes the variance reduction, and it reduces the size of samples with additional memory cost.
For the layer-wise sampling, Fast-GCN~\citep{DBLP:conf/iclr/ChenMX18} samples a fixed number of nodes at each layer, and ASGCN~\citep{DBLP:conf/nips/Huang0RH18} proposes the adaptive layer-wise sampling with better variance control.
In the graph level, Cluster-GCN~\citep{chiang2019cluster} firstly clusters the nodes and then samples the nodes in the clusters, and GraphSAINT~\citep{DBLP:conf/iclr/ZengZSKP20} directly samples a subgraph for mini-batch training. 
Recently, sampling has already been widely used in many GNNs and GNN systems~\citep{distdgl_ai3_2020, aligraph_vldb_2019, pygeometric_iclr_2019}.

\noindent\textbf{Graph-wise Propagation.} 
Recent studies have observed that non-linear feature transformation contributes little to the performance of the GNNs as compared to feature propagation.
Thus, a new direction for scalable GNN is based on the \emph{simplified} GCN (SGC)~\citep{wu2019simplifying}, which successively removes nonlinearities and collapsing weight matrices between consecutive layers. 
SGC reduces GNNs into a linear model operating on $K$-layers propagated features: 
\begin{equation}
\small
    \mathbf{X}^{(K)}=\mathbf{\hat{A}}^K \mathbf{X}^{(0)}, \qquad  \mathbf{Y} = \text{softmax}(\mathbf{\Theta} \mathbf{X}^{(K)}),
\end{equation}
\label{eq_GC}
\noindent\ignorespacesafterend
where $\mathbf{X}^{(0)} =\mathbf{X}$, $\mathbf{X}^{(K)}$ is the $K$-layers propagated feature, $\mathbf{\hat{A}} = \widetilde{\mathbf{D}}^{r-1}\widetilde{\mathbf{A}}\widetilde{\mathbf{D}}^{-r}$, and $\widetilde{\mathbf{A}}=\mathbf{A}+\mathbf{I}_{N}$ is the adjacency matrix $\mathbf{A}$ with self loops added.
$\mathbf{\hat{D}}$ is the corresponding degree matrix of $\mathbf{\hat{A}}$.
By setting $r = $ 0.5, 1 and 0, $\mathbf{\hat{A}}$ represents the symmetric normalization adjacency matrix  $\widetilde{\mathbf{D}}^{-1/2}\widetilde{\mathbf{A}}\widetilde{\mathbf{D}}^{-1/2}$~\citep{DBLP:conf/iclr/KlicperaBG19}, the transition probability matrix $\widetilde{\mathbf{A}}\widetilde{\mathbf{D}}^{-1}$~\citep{DBLP:conf/iclr/ZengZSKP20}, or the reverse transition probability matrix $\widetilde{\mathbf{D}}^{-1}\widetilde{\mathbf{A}}$~\citep{xu2018representation}, respectively. 
As the propagated features  $\mathbf{X}^{(K)}$ can be precomputed, SGC is easy to scale to large graphs. However, graph-wise propagation restricts the same propagation steps and a fixed RF for each node. Therefore, some nodes' features may be over-smoothed or under-smoothed due to the inconsistent RF expansion speed, leading to non-optimal performance.

\noindent\textbf{Layer-wise Propagation.} Following SGC, some recent methods adopt layer-wise propagation to combine the features with different propagation layers. SIGN~\citep{frasca2020sign} proposes to concatenate the propagated features at different propagation depth after simple linear transformation: $[\mathbf{X}^{(0)}\mathbf{W}_0, \mathbf{X}^{(1)}\mathbf{W}_1, ..., \mathbf{X}^{(K)}\mathbf{W}_K]$. S$^2$GC~\citep{zhu2021simple} proposes the simple spectral graph convolution to average the propagated features in different iterations as $\mathbf{X}^{(K)} = \sum \limits_{l=0}\limits^{K}\mathbf{\hat{A}}^l \mathbf{X}^{(0)}$. In addition, GBP~\citep{DBLP:conf/nips/ChenWDL00W20} further improves the combination process by weighted averaging as $\mathbf{X}^{(K)} = \sum \limits_{l=0}\limits^{K} w_l\mathbf{\hat{A}}^l \mathbf{X}^{(0)}$ with the layer weight $w_l = \beta {(1-\beta)}^l$.
We also use a linear model for higher training scalability similar to these works. 
The difference lies in that we consider the propagation process from a node-wise perspective, and each node in GAMLP has a personalized combination of different steps of the propagated features.

\subsection{Label Utilization in GNNs.}
Labels of training nodes are conventionally only used as supervision signals in loss functions in most graph learning methods.
However, there also exist some graph learning methods that directly exploit the labels of training nodes.
Among them, the label propagation algorithm~\citep{zhu2002learnin} is the most well-known one.
It simply regards the partially observed label matrix $\mathbf{Y} \in \mathbb{R}^{N \times C}$ as input features for nodes in the graph and propagates the input features through the graph structure, where $C$ is the number of candidate classes.
UniMP~\citep{shi2020masked} proposes to map the partially observed label matrix $\mathbf{Y}$ to the dimension of the node feature matrix $\mathbf{X}$ and add these two matrices together as the new input feature.
To fight against the label leakage problem, UniMP further randomly masks the training nodes during every training epoch.

Instead of using the hard training labels, Correct \& Smooth~\citep{huang2020combining} first trains a simple model (e.g., MLP) and gets the predicted soft labels for unlabeled nodes. Then, it propagates the learning errors on the labeled nodes to connected nodes and smooths the output in a Personalized PageRank manner like APPNP~\citep{DBLP:conf/iclr/KlicperaBG19}. 
Besides, SLE~\citep{sun2021scalable} decouples the label utilization procedure in UniMP, and executes the propagation in advance. 
Unlike UniMP, ``label reuse''~\citep{wang2021bag}  concatenates the partially observed label matrix $\mathbf{Y}$ with the node feature matrix $\mathbf{X}$ to form the new input matrix.
Concretely, it fills the missing elements in the partially observed label matrix $\mathbf{Y}$ with the soft label predicted by the model, and this newly generated $\mathbf{Y}'$ is again concatenated with $\mathbf{X}$ and then fed into the model.

\begin{figure*}[tpb]
    \centering
    \hspace{22.5mm}
    \includegraphics[width=0.85\textwidth]{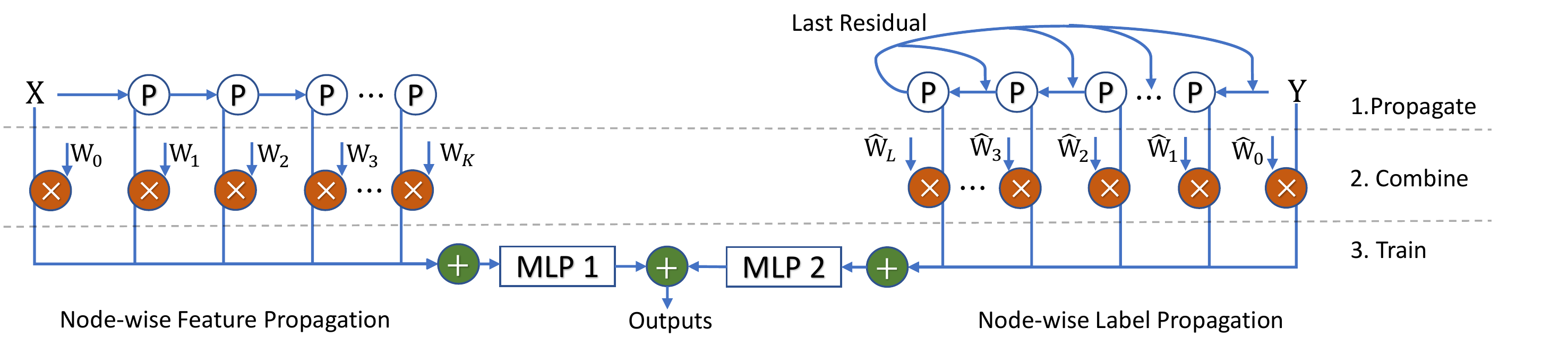}
    \vspace{-2mm}
    \caption{Overview of the proposed GAMLP, including (1) feature and label propagation, (2) combine the propagated features and labels with RF attention, and (3) MLP training. Note that both the feature and label propagation can be pre-processed.
}  
     \vspace{-2mm}
    \label{Fig.pipe}
\end{figure*}

\section{Graph Attention Multi-Layer Perceptron}
\label{sec4}
\subsection{Architecture Overview}
As shown in Fig.~\ref{Fig.pipe}, GAMLP decomposes the end-to-end GNN training into three parts: feature and label propagation, feature and label combination with RF attention, and the MLP training. 
As the feature and label propagation is pre-processed only once, and MLP training is efficient and salable, we can easily scale GAMLP to large graphs. Besides, with the RF attention, each node in GAMLP can adaptively get the suitable combination weights for propagated features and labels under different receptive fields, thus boosting model performance.

\subsection{Node-wise Feature and Label Propagation}  
\label{NFP}
\noindent\paragraph{\textbf{Node-wise Feature Propagation.}} We separate the essential operation of GNNs — feature propagation by removing the neural network $\mathbf{\Theta}$ and nonlinear activation $\delta$ for feature transformation. 
Specifically, we construct a parameter-free $K$-step feature propagation as:
\begin{equation}
\small
    \mathbf{X}^{(k)} \gets \hat{\mathbf{A}}\mathbf{X}^{(k-1)},\ \forall  k=1,\ldots, K,
    \label{equ:prop}
\end{equation}
where $\mathbf{X}^{(k)}$ contains the features of a fixed RF: the node itself and its $k$-hop neighborhoods. 

After $K$-step feature propagation shown in E.q.~\ref{equ:prop}, we correspondingly get a list of propagated features under different propagation steps: $[\mathbf{X}^{(0)}, \mathbf{X}^{(1)}, \mathbf{X}^{(k)}, ..., \mathbf{X}^{(K)}]$.
For a node-wise propagation, we propose to average these propagated features in a weighted manner:
\begin{equation}
\small
      \mathbf{H}_{\mathbf{X}}=\sum_{k=0}^K \mathbf{W}_k \mathbf{X}^{(k)} ,
      \label{combine}
\end{equation}
where $\mathbf{W}_k  = Diag(\eta_{k}) \in \mathbb{R}^{n \times n}$ is the diagonal matrix derived from vector $\eta_{k}$, and $\eta_{k} \in \mathbb{R}^{n}$ is a vector derived from vector $\eta_{k}[i] = w_{i}(k), 1\le i\le n$, and $w_{i}(k)$ measures the importance of the $k$-step propagated feature for node $v_i$.

\noindent\paragraph{\textbf{Node-wise Label Propagation.}} We use a scalable and node-adaptive way to take advantage of the node labels of the training set. 
Concretely, the label embedding matrix $\mathbf{Y} \in \mathbb{R}^{n \times c}$ ($\mathbf{Y}^{(0)}$) is propagated with the normalized adjacency matrix $\hat{\mathbf{A}}$:
\begin{small}
\begin{equation}
\small
    \mathbf{Y}^{(l)} \gets \hat{\mathbf{A}}\mathbf{Y}^{(l-1)},\ \forall  l=1,\ldots, L,
\end{equation}
\end{small}

After $L$-step label propagation, we get a list of propagated labels under different propagation steps: $[\mathbf{Y}^{(0)}, \mathbf{Y}^{(1)}, \mathbf{Y}^{(2)}, ..., \mathbf{Y}^{(L)}]$.
Generally, the propagated label $\mathbf{Y}^{(l)}$ is closer to the original label matrix $\mathbf{Y}^{(0)}$ with a smaller propagation step $l$, and thus face a higher risk of data leakage problem if it is directly used as the model input. We propose the last residual connection to adaptively smooth the different steps of propagated labels.
\begin{definition}[\textbf{Last Residual Connection}]
\label{df1}
Given the propagation step $l$, and a list of propagated labels: $[\mathbf{Y}^{(0)}, \mathbf{Y}^{(1)}, \mathbf{Y}^{(2)}, ..., \mathbf{Y}^{(L)}]$, we smooth each label $\mathbf{Y}^{(l)}$ with the smoothed label $\mathbf{Y}^{(L)}$:
\begin{equation}
\small
\label{iw}
\hat{\mathbf{Y}}^{(l)} \gets (1-\alpha_l)\mathbf{Y}^{(l)} + \alpha_l \mathbf{Y}^{(L)},\ l=1,\ldots, L,
\end{equation}
where $\alpha_l=\cos\left(\frac{\pi l}{2L}\right)$ controls the proportion of $\mathbf{Y}^{(L)}$ in the $l$-step propagated label.
\end{definition}

Similar to the node-wise feature propagation strategy  introduced in Sec.~\ref{NFP}, we propose to average these propagated labels in a weighted manner as follow:
\begin{equation}
\small
      \mathbf{H}_{\mathbf{Y}}=\sum_{l=0}^L \hat{\mathbf{W}}_l \hat{\mathbf{Y}}^{(l)}.
      \label{HY}
\end{equation}

\subsection{Node-adaptive Attention Mechanisms}
To satisfy different RF requirements for each node, we introduce two RF attention mechanisms to get $w_{i}(k)$.
Note that these attention mechanisms can be used in both the feature and label propagation, and we introduce them from a feature perspective here.
To apply them for node-wise label propagation, we only need to replace the feature $\mathbf{X}_i$ in Eq.~\ref{iw1} and Eq.~\ref{iw2} with the label $\mathbf{Y}_i$.
\label{RFA}

\begin{definition}[\textbf{Recursive Attention}]
\label{df2}
At each propagation step $l$, suppose $s \in \mathbb{R}^{d}$ is a learnable parameter vector, we recursively measure the feature information gain compared with the previous combined feature as:
\begin{equation}
\small
\label{iw1}
\widetilde{\mathbf{X}}_i^{(l)} = \mathbf{X}_i^{(l)} \parallel \sum_{k=0}^{l-1} w_i(k) \mathbf{X}_i^{(k)},  \quad  w_{i}(k) = {e^{\widetilde{w}_{i}(k)}}/{\sum \limits_{j=0}\limits^{l-1} e^{\widetilde{w}_{i}(j)}},
\end{equation}
\end{definition}
where $\parallel$ means concatenation, and ${\widetilde{w}_{i}(l)} = \delta(\widetilde{\mathbf{X}}_i^{(l)} \cdot s)$.
As $\widetilde{\mathbf{X}}_i^{(l-1)} \in \mathbb{R}^{d}$ combines the graph information under different propagation steps,  large proportion of the information in $\widetilde{\mathbf{X}}_i^{(l)}$ may have already existed in $\sum_{k=0}^{l-1} w_i(k) \mathbf{X}_i^{(k)}$, leading to small information gain.
A larger $w_{i}(l)$ indicates the feature $\mathbf{X}_i^{(l)}$ is more important to the current state of node $v_i$ since combining $\widetilde{\mathbf{X}}_i^{(l)}$ will introduce higher information gain.

Jumping Knowledge Network (JK-Net)~\citep{xu2018representation} adopts layer aggregation to combine the node embeddings of different GCN layers, and thus it can leverage the propagated nodes' information with different RF. Motivated by JK-Net, we propose to guide the feature combination process with the model prediction trained on all the propagated features.   
Concretely, GAMLP with JK attention includes two branches: the concatenated JK branch and the attention-based combination branch.
\begin{definition}[\textbf{JK Attention}]
\label{df3}
Given the MLP prediction of the JK branch as $ \mathbf{E}_i = \text{MLP}(\mathbf{X}_i^{(1)}\parallel\mathbf{X}_i^{(2)}\parallel ... \parallel\mathbf{X}_i^{(K)}) \in \mathbb{R}^{K \times f}$, the combination weight is defined as:
\begin{equation}
\small
\label{iw2}
\widetilde{\mathbf{X}}_i^{(l)} = \mathbf{X}_i^{(l)} \parallel \mathbf{E}_i, \quad    {\widetilde{w}_{i}(l)} = \delta(\widetilde{\mathbf{X}}_i^{(l)} \cdot s), \quad  w_{i}(l) = {e^{\widetilde{w}_{i}(l)}}/{\sum \limits_{k=0}\limits^{K} e^{\widetilde{w}_{i}(k)}}.
\end{equation}
\end{definition}
The JK branch aims to create a multi-scale feature representation for each node, which helps the attention mechanism learn the weight $w_i(k)$.
The learned weights are then fed into the attention-based combination branch to generate each node's refined attention feature representation.
As the training process continues, the attention-based combination branch will gradually 
emphasize those neighborhood regions that are more helpful to the target nodes.
The JK attention can model a broader neighborhood while enhancing correlations, bringing a better feature representation for each node.

\subsection{Model Training}
Both the combined feature $\mathbf{H}_{\mathbf{X}}$ and label $\mathbf{H}_{\mathbf{Y}}$ are transformed with MLP, and then be added to get the final output embedding:
\begin{equation}
\small
    \widetilde{\mathbf{H}} = \text{MLP}(\mathbf{H}_{\mathbf{X}}) + \beta\text{MLP}(\mathbf{H}_{\mathbf{Y}}),
\end{equation}
where $\beta$ is a hyper-parameter that measures the importance of the combined label. For example, some graphs have good features but low-quality labels (e.g., label noise or low label rate), and we should decrease $\beta$ so that more attention is paid to the graph features.

We adopt the Cross-Entropy (CE) measurement between the predicted softmax outputs and the one-hot ground-truth label distributions as the objective function: 
\begin{equation}
\small
\begin{aligned}
      & \mathcal{L}_{CE}= -\sum_{i \in \mathcal{V}_l } \sum_{j} \mathbf{Y}_{ij} \log(\text{softmax}(\widetilde{\mathbf{H}})_{ij}),
      \label{classi}
\end{aligned}
\end{equation}
where $\mathbf{Y}_i$ is the one-hot label indicator vector.

\begin{table*}[tpb]
\caption{Algorithm analysis for existing scalable GNNs.  $n$, $m$, $c$, and $f$ are the number of nodes, edges, classes,  and feature dimensions, respectively. $b$ is the batch size, and $k$ refers to the number of sampled nodes. $K$ and $L$ corresponds to the number of times we aggregate features and labels, respectively. Besides, $P$ and $Q$ are the number of layers in MLP classifiers trained with features and labels, respectively.} 
    \centering
    \resizebox{.75\linewidth}{!}{
    \begin{tabular}{c|c|c|c|c|c}
        \toprule
        \textbf{Type} & \textbf{Method} & \textbf{Pre-processing} & \textbf{Training}  & \textbf{Memory} \\
        \midrule
        \multirow{3}{*}{Sampling} & GraphSAGE & - & $\mathcal{O}(k^{K}nf^2)$  & $\mathcal{O}(bk^{K}f + Kf^2)$\\
        & FastGCN & - & $\mathcal{O}(kKnf^2)$  &  $\mathcal{O}(bkKf + Kf^2)$\\
        & Cluster-GCN & $\mathcal{O}(m)$ & $\mathcal{O}(Pmf+Pnf^2)$  & $\mathcal{O}(bKf + Kf^2)$\\
        \hline
        \multirow{1}{*}{Graph-wise propagation} & SGC & $\mathcal{O}(Kmf)$ & $\mathcal{O}(nf^2)$  &  $\mathcal{O}(bf+ f^2)$\\
        \hline
        \multirow{3}{*}{Layer-wise propagation}
        & SIGN & $\mathcal{O}(Kmf)$ & $\mathcal{O}(Pnf^2)$  &  $\mathcal{O}(bLf+ Pf^2)$\\
        & S$^2$GC & $\mathcal{O}(Kmf)$ & $\mathcal{O}(nf^2)$   &  $\mathcal{O}(bf+ f^2)$\\
        & GBP & $\mathcal{O}(Knf + K\frac{\sqrt{m\lg{n}}}{\varepsilon})$ & $\mathcal{O}(Pnf^2)$   & $\mathcal{O}(bf+ Pf^2)$\\
        \hline
        \multirow{1}{*}{Node-wise propagation}
        & GAMLP & $\mathcal{O}(Kmf+Lmc)$ & $\mathcal{O}(Pnf^2+Qnc^2)$   & $\mathcal{O}(bf+ Pf^2+ Qc^2)$\\
        % & MLP+LSL & $\mathcal{O}(LMc)$ & $\mathcal{O}(KNf^2)$   & $\mathcal{O}(KNf^2)$ & $\mathcal{O}(bf+ Kf^2)$\\
        \bottomrule
    \end{tabular}}
    
    \label{algorithm analysis}
\end{table*}

\subsection{Properties of \sys}
\noindent\textbf{High Efficiency and Scalability.}
Compared with the previous GNNs (e.g., GCN and GraphSAGE), our proposed GAMLP only need to do the feature and label propagation only once. Suppose $P$ and $Q$ are the number of layers in MLP trained with feature and labels, and $k$ is the sampled nodes, the time complexity of GAMLP is $\mathcal{O}(Pnf^2+Qnc^2)$, which is smaller than the complexity of GraphSAGE (i.e., $\mathcal{O}(k^Knf^2)$). Besides, it also costs less memory than the sampling-based GNNs, and thus can scale to a larger graph in a single machine. 
Notably, like other simplified GNNs (i.e., SGC and SIGN),  GAMLP can pre-compute the propagated features and labels only once. It does not need to pull the intermediate representation of other nodes during the MLP training. Therefore, it can also be well adapted to the distributed environment. 
% Further details can be found in Appendix~\ref{complexity}.

\noindent\textbf{Deep propagation.}
With our recursive and JK attention, GAMLP can support large propagation depths without the over-smoothing issue since each node can get the node personalized combination weights for different propagated features and labels according to its demand. Such characteristic is essential for sparse graph, i.e., sparse labels, edges, and features. For example, a graph with a low label rate or edge rate can increase the propagation depth to spread the label supervision over the entire graph. Each node can utilize the high-order graph structure information with deep propagation and boost the node classification performance. Further details are in Appendix~\ref{deep-pro}.

\subsection{Complexity Analysis}
\label{complexity}
Table~\ref{algorithm analysis} provides a detailed asymptotic complexity comparison between GAMLP and representative scalable GNN methods.
During preprocessing, the time cost of clustering in Cluster-GCN is $\mathcal{O}(m)$ and the time complexity of most linear models is $\mathcal{O}(Kmf)$. 
Besides, GAMLP has an extra time cost $\mathcal{O}(Lmc)$ for the propagation of training labels.
GBP takes advantage of Monte-Carlo method and conducts this process approximately with a bound of $\mathcal{O}(Knf + K\frac{\sqrt{m\lg{n}}}{\varepsilon})$, where $\varepsilon$ is a error threshold. 
Compared with sampling-based GNNs, graph/layer/node-wise-propagation-based models usually have smaller training and inference time complexity.
Memory complexity is a crucial factor in large-scale graph learning as it fundamentally determines whether it is possible to adopt the method.
Compared with SIGN, both GBP and GAMLP do not need to store smoothed features at different propagation steps, and the memory complexity can be reduced from $\mathcal{O}(bLf)$ to $\mathcal{O}(bf)$.

\section{Experiments}

In this section, we verify the effectiveness of GAMLP on 14 real-world graph datasets under (1) both the transductive and inductive settings; and (2) both the homogeneous and heterogeneous graphs.
We aim to answer the following five questions. 
\textbf{Q1:} Can GAMLP outperform the state-of-the-art GNN methods regarding predictive accuracy on real-world datasets?
\textbf{Q2:} If \sys is effective, where does the performance gain of GAMLP come from?
\textbf{Q3:} How does GAMLP perform when applied to highly sparse graphs (i.e., given few edges and low label rate)? 
\textbf{Q4:} Can GAMLP adapt to and perform well on heterogeneous graphs?
%\textbf{Q5:} How does GAMLP perform in real industrial production environment?
More experimental results can be found in Appendix~\ref{more-exp}.

\subsection{Experimental Setup}
\label{sec:settings}

\textbf{Datasets.}
We evaluate the performance of GAMLP under both transductive and inductive settings.
For transductive settings, we conduct experiments on 11 transductive datasets: three citation network datasets (Cora, Citeseer, PubMed)~\citep{DBLP:journals/aim/SenNBGGE08}, two user-item datasets (Amazon Computer, Amazon Photo), two co-author datasets (Coauthor CS, Coauthor Physics)~\citep{shchur2018pitfalls}, and three OGB datasets (ogbn-products, ogbn-papers100M, ogbn-mag)~\citep{hu2021ogb}, and one Tencent Video graph.
For inductive settings, we perform the comparison experiments on 3 widely used inductive datasets: PPI, Flickr, and Reddit~\citep{DBLP:conf/iclr/ZengZSKP20}.
The ogbn-mag dataset is also used to test the ability of \sys in a heterogeneous graph.
The statistics about these 14 datasets are summarized in Table~\ref{tab:data} of Appendix~\ref{app:dataset}.

\noindent\textbf{Baselines.} 
Under the transductive setting, we compare GAMLP with the following representative baseline methods: GCN~\citep{kipf2016semi}, GAT~\citep{velivckovic2017graph}, JK-Net~\citep{xu2018representation}, ResGCN~\citep{li2019deepgcns}, APPNP~\citep{DBLP:conf/iclr/KlicperaBG19}, AP-GCN~\citep{spinelli2020adaptive}, UniMP~\citep{shi2020masked}, SGC~\citep{wu2019simplifying}, SIGN~\citep{frasca2020sign}, S$^2$GC~\citep{zhu2021simple}, and GBP~\citep{DBLP:conf/nips/ChenWDL00W20}.
For the comparison in the OGB datasets, we choose the top-performing methods from the OGB leaderboard along with their accuracy results.
Under the inductive setting, we choose following representative methods: SGC~\citep{wu2019simplifying}, GraphSAGE~\citep{hamilton2017inductive}, Cluster-GCN~\citep{chiang2019cluster}, and GraphSAINT~\citep{DBLP:conf/iclr/ZengZSKP20}. 
Besides, for heterogeneous graph, we choose eight baseline methods from the OGB ogbn-mag leaderboard: R-GCN~\citep{schlichtkrull2018modeling}, SIGN~\citep{frasca2020sign}, HGT~\citep{hu2020heterogeneous}, R-GSN~\citep{wu2021r}, HGConv~\citep{yu2020hybrid}, R-HGNN~\citep{yu2021heterogeneous}, and NARS~\citep{yu2020scalable}. 

In addition, two variants of GAMLP are tested in the evaluation: GAMLP(JK) and GAMLP(R).
``JK'' and ``R'' stand for adopting ``JK attention'' and ``Recursive attention'' for the node-adaptive attention mechanism, respectively.
The detailed hyperparameters and experimental environment can be found in Appendix~\ref{hyperparameters} and Appendix~\ref{app:dataset}, respectively.

\begin{table}[tpb!]
\caption{Transductive performance on citation networks.} 
\label{tab:transductive}
% \vspace{-2mm}
\centering
{
\noindent
\renewcommand{\multirowsetup}{\centering}
\resizebox{0.75\linewidth}{!}{
\begin{tabular}{c|ccc}
\toprule
\textbf{Methods}&\textbf{Cora}& \textbf{Citeseer}&\textbf{PubMed}\\
\midrule
GCN& 81.8$\pm$0.5 & 70.8$\pm$0.5 &79.3$\pm$0.7\\
GAT& 83.0$\pm$0.7 & 72.5$\pm$0.7 &79.0$\pm$0.3\\
% \midrule
JK-Net& 81.8$\pm$0.5  & 70.7$\pm$0.7 & 78.8$\pm$0.7\\
ResGCN& 82.2$\pm$0.6 & 70.8$\pm$0.7 & 78.3$\pm$0.6\\
%\midrule
APPNP& 83.3$\pm$0.5 & 71.8$\pm$0.5 & 80.1$\pm$0.2\\
AP-GCN& 83.4$\pm$0.3& 71.3$\pm$0.5& 79.7$\pm$0.3\\
%\midrule
SGC & 81.0$\pm$0.2 & 71.3$\pm$0.5 & 78.9$\pm$0.5\\
SIGN & 82.1$\pm$0.3 & 72.4$\pm$0.8 &79.5$\pm$0.5\\
S$^2$GC & 82.7$\pm$0.3 & 73.0$\pm$0.2 & 79.9$\pm$0.3 \\
GBP & \underline{83.9$\pm$0.7} & 72.9$\pm$0.5 & 80.6$\pm$0.4\\
\midrule
\textbf{GAMLP(JK)} & \textbf{84.3$\pm$0.8} & \textbf{74.6$\pm$0.4} & \underline{80.7$\pm$0.4} \\ 
\textbf{GAMLP(R)} & \underline{83.9$\pm$0.6} & \underline{73.9$\pm$0.6} & \textbf{80.8$\pm$0.5}\\
\bottomrule
\end{tabular}}}
\vspace{-3mm}
\end{table}

\begin{table}[tpb!]
\caption{Transductive performance on the co-authorship and co-purchase graphs.}
\label{tab:transductive2}
% \vspace{-2mm}
\centering
{
\noindent
\renewcommand{\multirowsetup}{\centering}
\resizebox{0.85\linewidth}{!}{
\begin{tabular}{c|cccc}
\toprule
\textbf{Methods}&{\textbf{\makecell{Amazon \\Computer}}} & 
{\textbf{\makecell{Amazon \\Photo}}} & 
{\textbf{\makecell{Coauthor \\CS}}}&
{\textbf{\makecell{Coauthor \\Physics}}} \\
\midrule
GCN&82.4$\pm$0.4 & 91.2$\pm$0.6 & 90.7$\pm$0.2 & 92.7$\pm$1.1\\
GAT& 80.1$\pm$0.6& 90.8$\pm$1.0 & 87.4$\pm$0.2 & 90.2$\pm$1.4\\
% \midrule
JK-Net& 82.0$\pm$0.6 & 91.9$\pm$0.7 & 89.5$\pm$0.6 & 92.5$\pm$0.4\\
ResGCN&81.1$\pm$0.7 & 91.3$\pm$0.9 & 87.9$\pm$0.6 & 92.2$\pm$1.5\\
%\midrule
APPNP&81.7$\pm$0.3&91.4$\pm$0.3&92.1$\pm$0.4&92.8$\pm$0.9\\
AP-GCN&83.7$\pm$0.6& 92.1$\pm$0.3& 91.6$\pm$0.7& 93.1$\pm$0.9\\
%\midrule
SGC &82.2$\pm$0.9&91.6$\pm$0.7&90.3$\pm$0.5& 91.7$\pm$1.1\\
SIGN &83.1$\pm$0.8&91.7$\pm$0.7&91.9$\pm$0.3& 92.8$\pm$0.8\\
S$^2$GC & 83.1$\pm$0.7 & 91.6$\pm$0.6 & 91.6$\pm$0.6 & 93.1$\pm$0.8 \\
GBP & 83.5$\pm$0.8 & 92.1$\pm$0.8 & 92.3$\pm$0.4 & \underline{93.3$\pm$0.7}\\
\midrule
\textbf{GAMLP(JK)}& \textbf{84.5$\pm$0.7} & \textbf{92.8$\pm$0.7} & \underline{92.6$\pm$0.5} & \textbf{93.6$\pm$1.0} \\ 
\textbf{GAMLP(R)} & \underline{84.2$\pm$0.5} & \underline{92.6$\pm$0.8} & \textbf{92.8$\pm$0.7} & 93.2$\pm$1.0 \\
\bottomrule
\end{tabular}}}
\vspace{-3mm}
\end{table}

\begin{table}[tpb!]
     \centering
     \captionof{table}{Performance comparison on ogbn-products.} 
    \label{tab:products_perf}
     \begin{tabular}{c|cc}
    \toprule
    \textbf{Methods} & \textbf{Val Accuracy} & \textbf{Test Accuracy} \\
    \midrule
    GCN & 92.00$\pm$0.03 & 75.64$\pm$0.21 \\
    SGC & 92.13$\pm$0.02 & 75.87$\pm$0.14 \\
    GraphSAGE & 92.24$\pm$0.07 & 78.50$\pm$0.14 \\
    GraphSAINT & 92.52$\pm$0.13 & 80.27$\pm$0.26 \\
    GBP & 92.82$\pm$0.10 & 80.48$\pm$0.05 \\       
    SIGN & 92.99$\pm$0.04 & 80.52$\pm$0.16 \\
    DeeperGCN & 92.38$\pm$0.09 & 80.98$\pm$0.20 \\
    UniMP & 93.08$\pm$0.17 & 82.56$\pm$0.31 \\
    SAGN & 93.09$\pm$0.04 & 81.20$\pm$0.07 \\
    SAGN+0-SLE & 93.27$\pm$0.04 & 83.29$\pm$0.18\\
    \midrule
    \textbf{GAMLP(JK)} & 93.19$\pm$0.03 & \underline{83.54$\pm$0.25}  \\
    \textbf{GAMLP(R)} & 93.11$\pm$0.05 & \textbf{83.59$\pm$0.09}  \\
    \bottomrule
    \end{tabular}
    \vspace{-3mm}
\end{table}

\begin{table}[tpb!]
 \centering
 \captionof{table}{Performance comparison on ogbn-papers100M.} 
\label{tab:100m_perf}
 \begin{tabular}{c|cc}
\toprule
\textbf{Methods} & \textbf{Val Accuracy} & \textbf{Test Accuracy} \\
\midrule
SGC & 66.48$\pm$0.20 & 63.29$\pm$0.19 \\
SIGN & 69.32$\pm$0.06 & 65.68$\pm$0.06 \\
SIGN-XL & 69.84$\pm$0.06 & 66.06$\pm$0.19 \\
SAGN & 70.34$\pm$0.99 & 66.75$\pm$0.84 \\
SAGN+0-SLE & 71.06$\pm$0.08 & \underline{67.55$\pm$0.15}\\
\midrule
\textbf{GAMLP(JK)} & 71.92$\pm$0.04 & \textbf{68.07$\pm$0.10}  \\
\textbf{GAMLP(R)} & 71.21$\pm$0.03 & 67.46$\pm$0.02  \\
\bottomrule
\end{tabular}
\vspace{-3mm}
\end{table}

\subsection{End-to-end Comparison}  

\textbf{Transductive Performance.}  To answer \textbf{Q1}, we report the transductive performance of GAMLP in Tables~\ref{tab:transductive},~\ref{tab:transductive2}~\ref{tab:products_perf}, and~\ref{tab:100m_perf}. 
We observe that both variants of GAMLP outperform all the baseline methods on almost all the datasets.
For example, on the small Citeseer dataset, GAMLP(JK) outperforms the state-of-the-art method S$^2$GC by a large margin of 1.6\%; on the medium-sized Amazon Computers, the predictive accuracy of GAMLP (JK) exceeds the one of the state-of-the-art method GBP by 1.0\%; on the two large OGB datasets, GAMLP takes the lead by 1.03\% and 1.32\% on ogbn-products and ogbn-papers100M, respectively.
Furthermore, the experimental results illustrate that the contest between the two variants of GAMLP is not a one-horse race.
Thus, these two different attention mechanisms have their irreplaceable sense in some ways.

\begin{table}
    \begin{center}
    % \vspace{-5mm}
    \caption{Performance comparison on three inductive datasets.}
    \label{tab:inductive}
    \vspace{-2mm}
    \begin{tabular}{c|ccc}
    \toprule
    \textbf{Methods} & \textbf{PPI} & \textbf{Flickr} & \textbf{Reddit} \\
    \midrule
    SGC & 65.7$\pm$0.01 & 50.2$\pm$0.12 & 94.9$\pm$0.00 \\
    GraphSAGE &  61.2$\pm$0.05  &  50.1$\pm$0.13  &  95.4$\pm$0.01   \\
    Cluster-GCN &  99.2$\pm$0.04  &  48.1$\pm$0.05  &  95.7$\pm$0.00  \\
    GraphSAINT& 99.4$\pm$0.03 & 51.1$\pm$0.10 & \underline{96.6$\pm$0.01} \\
    \midrule
    \textbf{GAMLP(JK)}   & \textbf{99.82$\pm$0.01} & \textbf{54.12$\pm$0.01} & \textbf{97.04$\pm$0.01} \\    
    \textbf{GAMLP(R)}   & \underline{99.66$\pm$0.01} & \underline{53.12$\pm$0.00} & \underline{96.62$\pm$0.01} \\       
    \bottomrule
    \end{tabular}
    \end{center}
    \vspace{-3mm}
\end{table}

\noindent\textbf{Inductive Performance.} 
The experiment results in Table~\ref{tab:inductive} show that GAMLP consistently outperforms all the baseline methods under the inductive setting.
The leading gap of GAMLP(JK) over SOTA inductive method -- GraphSAINT is more than 3.0\% on the widely-used dataset -- Filckr.
The impressive performance of GAMLP under the inductive setting illustrates that GAMLP is powerful in predicting the properties of unseen nodes.

\subsection{Ablation Study}
To answer \textbf{Q2}, we focus on two modules in GAMLP: (1) label utilization; (2) attention mechanism in the node-wise propagation.
For the second one, we evaluate the effects of different choices for reference vectors in the JK attention.

\noindent\textbf{Label Utilization.}
In this part, we evaluate whether adding the last residual connection and making use of training labels help or not.
The predictive accuracy of GAMLP(R) is evaluated on the ogbn-products dataset along with its three variants: ``-no\_label'', ``-plain\_label'', and ``-uniform'', which stands for not using labels, removing last residual connections, and replacing last residual connections with uniform distributions, respectively.
The experimental results in Table~\ref{tab:ablation_label} show that utilizing labels brings significant performance gain to GAMLP: from 81.43\% to 83.59\%.
The performance drop from removing the last residual connections (``-plain\_label'' in Table~\ref{tab:ablation_label}) is significant since directly adopting the raw training labels leads to the overfitting issue.
The fact that ``-uniform'' performs worse than ``-no\_label'' illustrates that intuitively fusing the original label distribution with the uniform distribution would harm the predictive accuracy.
It further demonstrates the effectiveness of our proposed last residual connections.

\noindent\textbf{Reference Vector in Attention Mechanism.} In this part, we study the role of the reference vector (set originally as the concatenated features from different propagation steps) in our proposed JK attention.
We evaluate the three variants of GAMLP(JK): ``-origin\_feature'', ``-normal\_noise'', and ``-no\_reference'', which changes the reference vector to the original node feature, noise from the normal distribution, and nothing, respectively.
The predictive accuracy of each variant on the PubMed dataset is reported in Table~\ref{tab:ablation_reference}.
The experimental results show that our original choice of the reference vector is the best among itself and its three variants.
The superiority of the concatenated features from different propagation steps comes from the fact that it allows the model to capture the interactions between the propagated features over the receptive fields with different sizes.

\begin{table}
     \centering
     \captionof{table}{Ablation study on label utilization.}
    \label{tab:ablation_label}
     \begin{tabular}{c|cc}
\toprule
\textbf{Methods} & \textbf{Val Accuracy} & \textbf{Test Accuracy} \\
\midrule
GAMLP(R) & 93.11$\pm$0.05 & \textbf{83.59$\pm$0.05}  \\
-no\_label & 92.29$\pm$0.06 & 81.43$\pm$0.18  \\
-plain\_label & 92.53$\pm$0.21 & 81.12$\pm$0.45  \\
-uniform &92.72$\pm$0.15 & 81.28$\pm$0.93  \\
\bottomrule
\end{tabular}
\end{table}

\begin{table}
     \centering
    \captionof{table}{Ablation study on reference vector.}
    \label{tab:ablation_reference}
     \begin{tabular}{c|cc}
\toprule
\textbf{Methods} & \textbf{Val Accuracy} & \textbf{Test Accuracy} \\
\midrule
GAMLP(JK) & 82.5$\pm$0.5 &  \textbf{80.7$\pm$0.4}\\
-origin\_feature & 82.2$\pm$0.4 & 80.5$\pm$0.4  \\
-normal\_noise & 81.8$\pm$0.4 & 79.8$\pm$0.5 \\
-no\_reference & 81.5$\pm$0.5 & 79.9$\pm$0.3  \\
\bottomrule
\end{tabular}
% \vspace{-3mm}
\end{table}

\begin{table}[tpb!]
\caption{Test accuracy on ogbn-mag dataset.}
% \vspace{-2mm}
\centering
{
\noindent
\renewcommand{\multirowsetup}{\centering}
\resizebox{0.75\linewidth}{!}{
\begin{tabular}{c|cc}
\toprule
\textbf{Methods} & \textbf{Validation Accuracy} & \textbf{Test Accuracy} \\
\midrule
R-GCN & 40.84$\pm$0.41 & 39.77$\pm$0.46 \\
SIGN & 40.68$\pm$0.10 & 40.46$\pm$0.12 \\
HGT & 49.84$\pm$0.47 & 49.27$\pm$0.61 \\
R-GSN & 51.82$\pm$0.41 & 50.32$\pm$0.37 \\
HGConv & 53.00$\pm$0.18 & 50.45$\pm$0.17 \\
R-HGNN & 53.61$\pm$0.22 & 52.04$\pm$0.26 \\
NARS & 53.72$\pm$0.09 & \underline{52.40$\pm$0.16} \\
\textbf{NARS-GAMLP} & 55.52$\pm$0.08 & \textbf{54.01$\pm$0.21}  \\
\bottomrule
\end{tabular}}}
\label{table.mag_performance}
\vspace{-4mm}
\end{table}

\subsection{Performance on Sparse Graphs}
\begin{figure}[tp]
\centering  
% \vspace{-4mm}
\subfigure[Edge Sparsity]{
\includegraphics[width=0.23\textwidth]{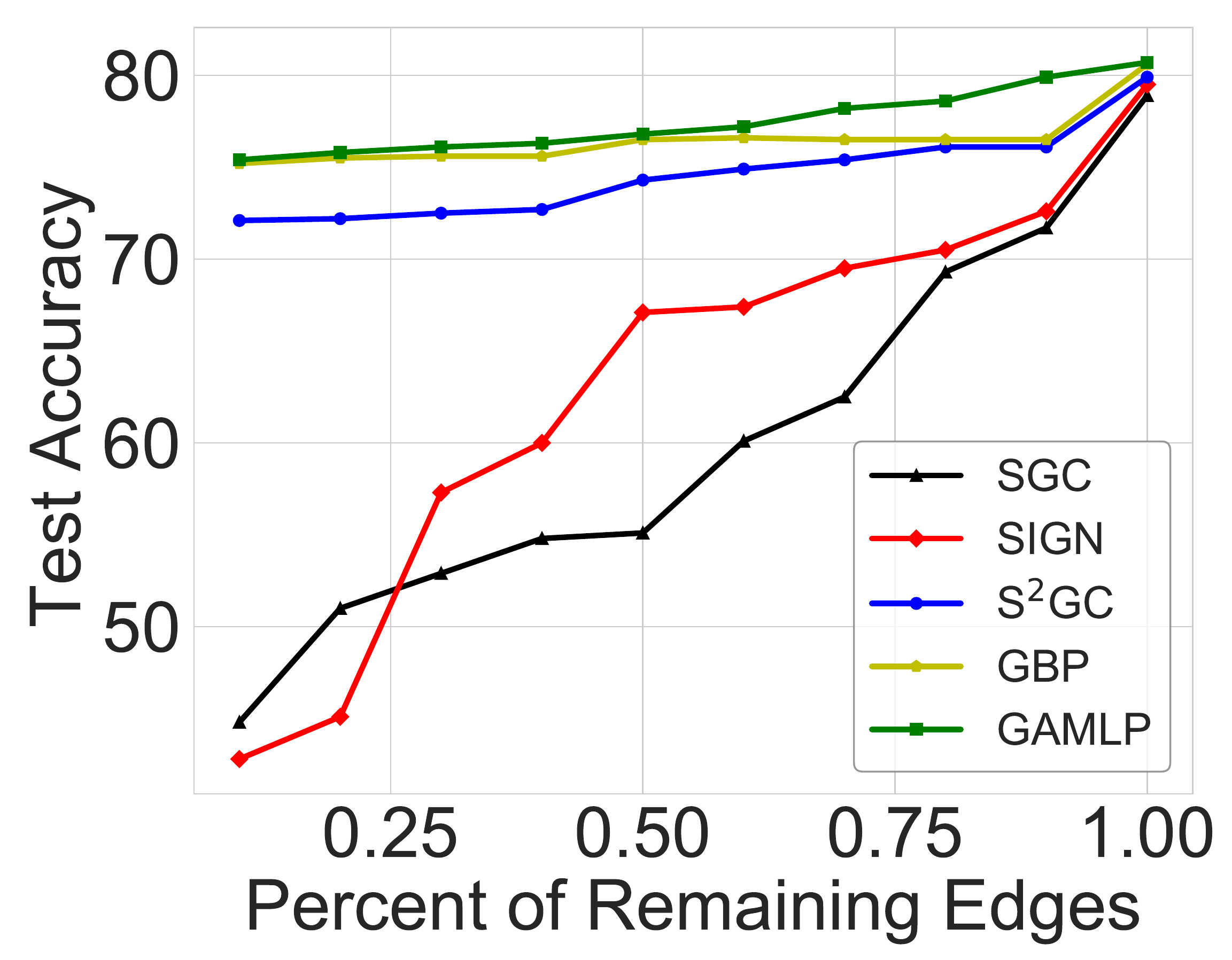}}
\subfigure[Label Sparsity]{
\includegraphics[width=0.23\textwidth]{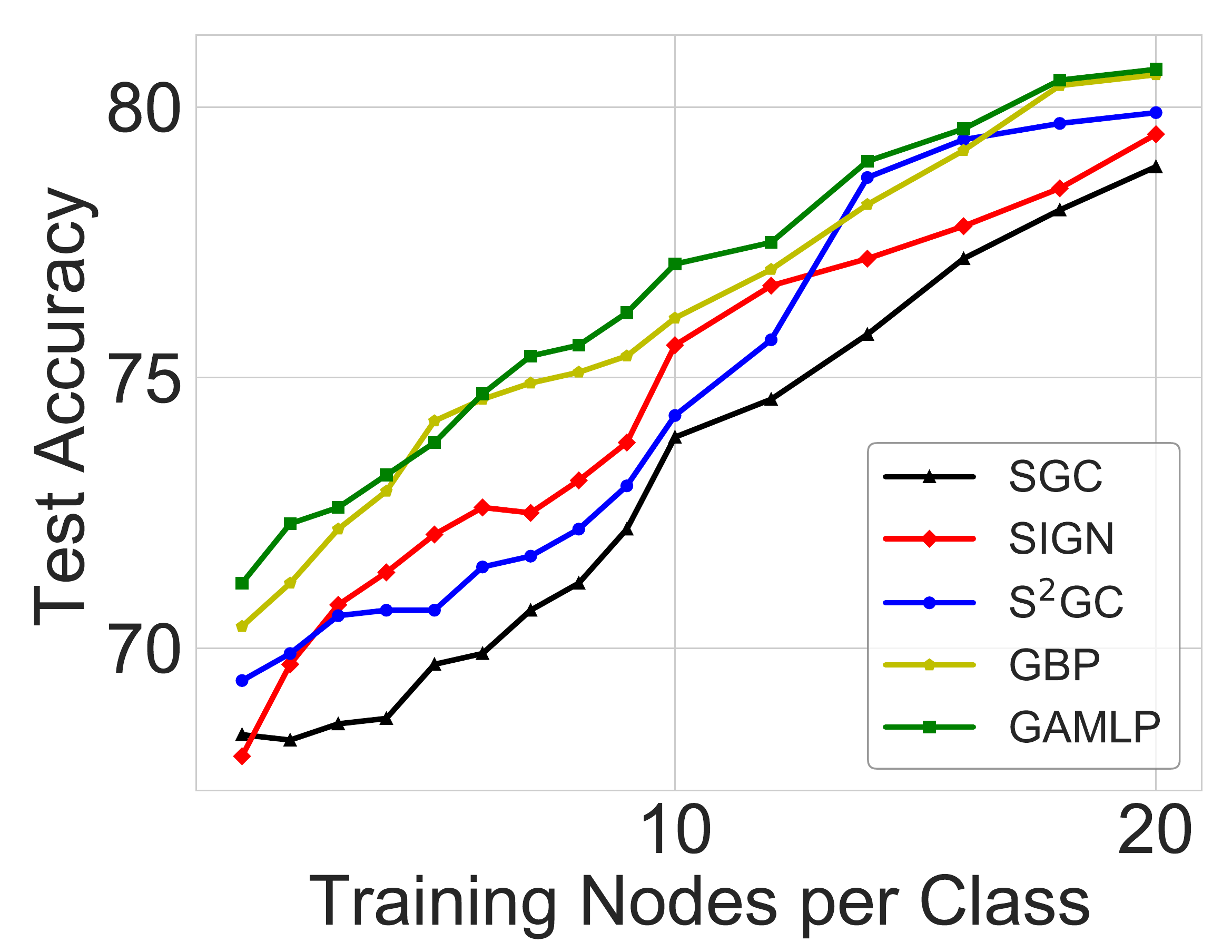}}
\vspace{-2mm}
\caption{Test accuracy on PubMed dataset under different levels of label and edge sparsity.}
\label{fig.sparsity}
\vspace{-4mm}
\end{figure}

To answer \textbf{Q3}, we conduct experiments to evaluate the predictive accuracy of GAMLP when faced with edge and label sparsity problems, where the number of edges and training labels are highly scarce.
We randomly remove a fixed percentage of edges from the original graph to simulate the edge sparsity problem.
The removed edges are precisely the same for all the compared methods.
Besides, we enumerate the number of training nodes per class from 1 to 20 to evaluate the performance of GAMLP given different levels of label sparsity.
The experimental results in Fig.~\ref{fig.sparsity} show that GAMLP outperforms all the baselines in most cases when faced with different levels of edge and label sparsity.
This experiment further demonstrates the effectiveness of our proposed node-wise propagation scheme.
The node-wise propagation enables GAMLP to better capture long-range dependencies, which is crucial when applying GNN methods to highly sparse graphs.

\subsection{Performance on Heterogeneous Graphs}
Heterogeneous graphs are widely used in many real-world applications. Thus, we measure the performance of GAMLP in heterogeneous graphs and answer~\textbf{Q4}. Note that the ogbn-mag dataset only contains node features for ``paper'' nodes, and we here adopt the ComplEx algorithm~\citep{trouillon2017knowledge} to generate features for other nodes.
 
\noindent\textbf{Adapting GAMLP to Heterogeneous Graphs.}
We follow the design of NARS~\cite{yu2020scalable} to adapt GAMLP to heterogeneous graphs.
First, we sample subgraphs from the original heterogeneous graphs according to the different edge type combinations and regard the sampled subgraph as a homogeneous graph.
Then, the propagated features of different steps are generated on each subgraph.
The propagated features and labels of the same propagation step across different subgraphs are aggregated using 1-d convolution.
After that, aggregated features and labels of different steps are fed into our GAMLP to get the final results.
This variant of our GAMLP is called NARS-GAMLP, and we adopt the ``JK attention'' here.

\noindent\textbf{Experimental Results.}
We report the validation and test accuracy of our proposed NARS-GAMLP on the ogbn-mag dataset in Table~\ref{table.mag_performance}.
It can be seen from the results that NARS-GAMLP achieves state-of-the-art performance on the heterogeneous graph dataset ogbn-mag.
Specifically, it outperforms the strongest single model baseline NARS by a large margin of 1.61\% regarding test accuracy.

\begin{table*}[tbp!]
\vspace{-2mm}
\caption{Efficiency and accuracy comparison on the Tencent video classification.}
\vspace{-2mm}
\centering
{
\noindent
\renewcommand{\multirowsetup}{\centering}
\resizebox{.95\linewidth}{!}{
\begin{tabular}{c|cccccccccccc}
\toprule
&\textbf{SGC}& \textbf{S$^2$GC}&\textbf{GBP}&\textbf{SIGN}&\textbf{GAMLP(R)}&\textbf{\sys(JK)}&\textbf{GCN}&\textbf{APPNP}&\textbf{AP-GCN}&\textbf{JK-Net}&\textbf{ResGCN}&\textbf{GAT}\\
\midrule
Training Time & 1.0 & 1.2 & 1.3 & 3.2 & 6.1 & 7.4 & 33.1 & 77.8 & 112.3 & 112.8 & 132.3 & 372.4\\
Test Accuracy & 45.2$\pm$0.3 & 46.6$\pm$0.6 & 46.9$\pm$0.7 & 46.3$\pm$0.5 & \underline{47.8}$\pm$0.4& \textbf{48.1}$\pm$0.6  & 45.9$\pm$0.4 & 46.7$\pm$0.6& 46.9$\pm$0.7& 47.2$\pm$0.3  & 45.8$\pm$0.5  & 46.8$\pm$0.7\\
\bottomrule
\end{tabular}}}
%\vspace{-2mm}
\label{tab:efficiency}
% \vspace{-3mm}
\end{table*}

\begin{figure*}[tpb]
    \centering
    % \vspace{-2mm}
    \includegraphics[width=0.9\textwidth]{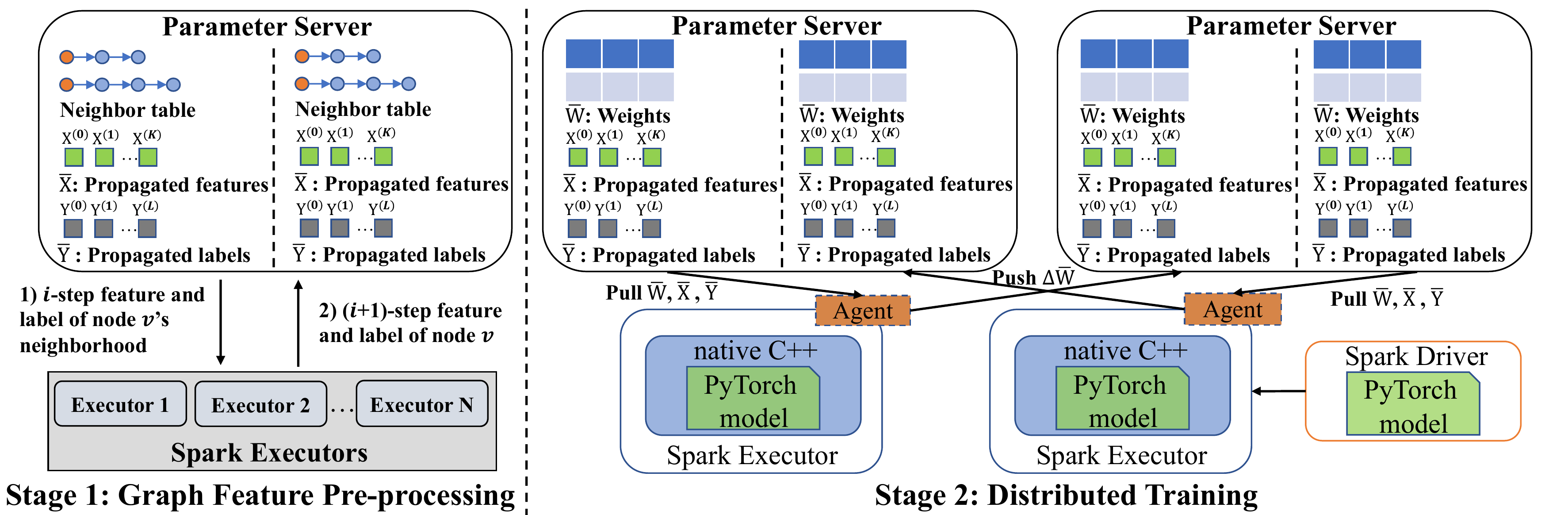}
    \vspace{-1mm}
    \caption{An overview of \sys deployed in Tencent.
}  
    \vspace{-1mm}
    \label{Fig.implementation}
\end{figure*}

\section{Deployment in Tencent}
We now introduce the implementation and deployment
of \sys in Tencent --- the largest social media conglomerate in China. 

\subsection{\sys Training Framework}
Unlike most GNNs that entangle the propagation and transformation in the training process, the training of \sys is separated into two stages: the graph feature pre-processing and the distributed model training.  
First, we pre-compute the propagated features and labels with different propagation steps, and then we train the model parameters with the parameter server. 
Note that both these two stages are implemented in a distributed fashion, and the implementation details of \sys can be found in Fig.~\ref{Fig.implementation}.

\noindent\textbf{Graph feature pre-processing.}
For the first pre-processing stage, we store the neighbor table, the propagated node features, and labels in a distributed manner. We recursively calculate the propagated features $\overline{\mathbf{X}} = [\mathbf{X}^{(0)},\mathbf{X}^{(1)} ..., \mathbf{X}^{(K)}]$ and the propagated labels $\overline{\mathbf{Y}} = [\mathbf{Y}^{(0)},\mathbf{Y}^{(1)} ..., \mathbf{Y}^{(L)}]$ for better efficiency. Specifically, to calculate the $i$-th step feature and label for each node $v$ in a batch, we firstly pull the corresponding ($i-1$)-th step feature and label information of its neighbors and then push the propagated feature back to the distributed storage for the calculation of ($i+1$)-th step. The computation of feature and label propagation in each batch is implemented by the Spark executors~\cite{zaharia2010spark} in parallel with matrix multiplication.
Since we compute the propagated features and labels in parallel, the graph feature pre-processing in the implementation of \sys could scale to large graphs and significantly improve the training efficiency in real-world applications.

\noindent\textbf{Distributed training.}
We implement \sys by Angel~\cite{jiang2020psgraph} for the second stage and optimize the parameters with distributed SGD. Specifically, the Spark executors frequently pull the model weights $\overline{\mathbf{W}}$ (including both the attention matrix for weighted average and the parameters of the MLP model), the propagated features $\overline{\mathbf{X}}$ and labels $\overline{\mathbf{Y}}$ from the parameter server.
Unlike existing GNN systems 
(e.g., DistDGL~\cite{distdgl_ai3_2020} and FlexGraph~\cite{wang2021flexgraph}) that need to sample and pull the neighborhood features in each training epoch, the neighbor table is not used in the training process of \sys, and we treat each node as independent and identically distributed. Therefore, the communication cost of training \sys can be significantly reduced. Since each spark executor can independently fetches the most up-to-date model weights $\overline{\mathbf{W}}$ and update them in a distributed manner, \sys can easily scale to large graphs.

\noindent\textbf{Workflow.}
The workflow of \sys can be summarized as the following steps: 1) The users write the PyTorch scripts and get the PyTorch model. 2) The Spark executor read the PyTorch model and datasets, and pre-computes the propagated features and labels in a distributed manner. 3) The Spark driver pushes the initialized PyTorch model to the parameter server (PS). 4) At each training epoch, every executor samples a batch of nodes and pulls the corresponding model weights, the propagated features, and labels from PS. Then, it updates the model weights with the backpropagation, and pushes the gradients back to PS. Note that we embed PyTorch inside Spark, and we transfer data between JVM runtime and C++
runtime using JNI (Java Native Interface). With our implementation, the users can simply write the python scripts in PyTorch while benefiting from Spark's distributed data processing.

\subsection{Results in WeSee Video Classification}
\sys has provided service to many applications in Tencent, such as WeSee~\footnote{\blue{\url{{https://weishi.qq.com/}}}} short-video classification and WeChat payment prediction. Here we show the effectiveness and efficiency of \sys on the short-video classification for the WeSee, a TikTok-like short-video service of Tencent. 
This task aims to classify short videos into pre-defined 253 classes related to different modalities (such as emotions, theme, place, et al.), which is an important prerequisite for video content understanding and recommendation in WeSee.

\noindent\textbf{Graph construction.} We collect 1,000,000 short-videos from the Wesee APP with 57,022 of them manually labeled and then generate a bipartite user-video graph. 
The edge between each node pair represents that the user has watched (clicked) the short video. 
Besides, we select 5,000 nodes as the train set, 30,000 nodes as the test set, and an additional validation set of 10,000 labeled nodes for hyper-parameter tuning.

\noindent\textbf{Classification results.} Table~\ref{tab:efficiency} illustrates the relative training time of each compared method along with its predictive accuracy.
The training time of SGC is set to $1.0$ as reference.
We observe that (1) The graph/layer-wise propagation-based methods (e.g., SGC and SIGN) have a significant advantage over many widely used GNNs (e.g., GCN and GAT) regarding training efficiency.
(2) The two variants of GAMLP achieve the highest predictive accuracy while the training time is acceptable compared to more demanding methods like JK-Net and AP-GCN.
The ``cold'' short videos are less popular, and the corresponding nodes lie in a sparse region of the graph. Thus they need more propagation steps to enhance their node embedding with their distant neighbors. However, simply adopting a large propagation depth will make the propagated node embedding for some ``hot'' short videos watched by most users indistinguishable.
In such cases, the node-wise feature and label propagation of \sys are essential to preserving the personalized information of short videos.

\section{Conclusion}
We present Graph Attention Multilayer Perceptron (GAMLP), a scalable, efficient, and deep graph model based on receptive field attention. 
GAMLP introduces two new attention mechanisms: recursive attention and JK attention, enabling learning the representations over RF with different sizes in a node-adaptive manner. We have deployed \sys in Tencent, and it has served many real-world applications.
Extensive experiments on 14 graph datasets verified the high predictive performance of \sys. Specifically, in the large-scale short-video dataset from the WeSee APP, the proposed \sys exceeded the compared baselines by a large margin in test accuracy while achieving comparable training time with SGC.
\sys moves forward the performance boundary of scalable GNNs, especially on large-scale and sparse industrial graphs.

\begin{acks}
This work is supported by NSFC (No. 61832001, 61972004), Beijing Academy of Artificial Intelligence (BAAI), and PKU-Tencent Joint Research Lab. Bin Cui is the corresponding author.
\end{acks}

\bibliographystyle{ACM-Reference-Format}
\bibliography{sample-base}


\begin{thebibliography}{47}

%%% the Web address.

\ifx \showCODEN    \undefined \def \showCODEN     #1{\unskip}     \fi
\ifx \showDOI      \undefined \def \showDOI       #1{#1}\fi
\ifx \showISBNx    \undefined \def \showISBNx     #1{\unskip}     \fi
\ifx \showISBNxiii \undefined \def \showISBNxiii  #1{\unskip}     \fi
\ifx \showISSN     \undefined \def \showISSN      #1{\unskip}     \fi
\ifx \showLCCN     \undefined \def \showLCCN      #1{\unskip}     \fi
\ifx \shownote     \undefined \def \shownote      #1{#1}          \fi
\ifx \showarticletitle \undefined \def \showarticletitle #1{#1}   \fi
\ifx \showURL      \undefined \def \showURL       {\relax}        \fi
% invisible to TeX
\providecommand\bibfield[2]{#2}
\providecommand\bibinfo[2]{#2}
\providecommand\natexlab[1]{#1}
\providecommand\showeprint[2][]{arXiv:#2}

\bibitem[Chen et~al\mbox{.}(2018a)]%
        {DBLP:conf/iclr/ChenMX18}
\bibfield{author}{\bibinfo{person}{Jie Chen}, \bibinfo{person}{Tengfei Ma},
  {and} \bibinfo{person}{Cao Xiao}.} \bibinfo{year}{2018}\natexlab{a}.
\newblock \showarticletitle{FastGCN: Fast Learning with Graph Convolutional
  Networks via Importance Sampling}. In \bibinfo{booktitle}{\emph{ICLR}}.
\newblock


\bibitem[Chen et~al\mbox{.}(2018b)]%
        {DBLP:conf/icml/ChenZS18}
\bibfield{author}{\bibinfo{person}{Jianfei Chen}, \bibinfo{person}{Jun Zhu},
  {and} \bibinfo{person}{Le Song}.} \bibinfo{year}{2018}\natexlab{b}.
\newblock \showarticletitle{Stochastic Training of Graph Convolutional Networks
  with Variance Reduction}. In \bibinfo{booktitle}{\emph{ICML}}.
  \bibinfo{pages}{941--949}.
\newblock


\bibitem[Chen et~al\mbox{.}(2020)]%
        {DBLP:conf/nips/ChenWDL00W20}
\bibfield{author}{\bibinfo{person}{Ming Chen}, \bibinfo{person}{Zhewei Wei},
  \bibinfo{person}{Bolin Ding}, \bibinfo{person}{Yaliang Li},
  \bibinfo{person}{Ye Yuan}, \bibinfo{person}{Xiaoyong Du}, {and}
  \bibinfo{person}{Ji{-}Rong Wen}.} \bibinfo{year}{2020}\natexlab{}.
\newblock \showarticletitle{Scalable Graph Neural Networks via Bidirectional
  Propagation}. In \bibinfo{booktitle}{\emph{NeurIPS}}.
\newblock


\bibitem[Chiang et~al\mbox{.}(2019)]%
        {chiang2019cluster}
\bibfield{author}{\bibinfo{person}{Wei-Lin Chiang}, \bibinfo{person}{Xuanqing
  Liu}, \bibinfo{person}{Si Si}, \bibinfo{person}{Yang Li},
  \bibinfo{person}{Samy Bengio}, {and} \bibinfo{person}{Cho-Jui Hsieh}.}
  \bibinfo{year}{2019}\natexlab{}.
\newblock \showarticletitle{Cluster-gcn: An efficient algorithm for training
  deep and large graph convolutional networks}. In
  \bibinfo{booktitle}{\emph{SIGKDD}}. \bibinfo{pages}{257--266}.
\newblock


\bibitem[Cui et~al\mbox{.}(2020)]%
        {cui2020adaptive}
\bibfield{author}{\bibinfo{person}{Ganqu Cui}, \bibinfo{person}{Jie Zhou},
  \bibinfo{person}{Cheng Yang}, {and} \bibinfo{person}{Zhiyuan Liu}.}
  \bibinfo{year}{2020}\natexlab{}.
\newblock \showarticletitle{Adaptive Graph Encoder for Attributed Graph
  Embedding}. In \bibinfo{booktitle}{\emph{SIGKDD}}. \bibinfo{pages}{976--985}.
\newblock


\bibitem[Fey and Lenssen(2019)]%
        {pygeometric_iclr_2019}
\bibfield{author}{\bibinfo{person}{Matthias Fey} {and} \bibinfo{person}{Jan~E.
  Lenssen}.} \bibinfo{year}{2019}\natexlab{}.
\newblock \showarticletitle{Fast Graph Representation Learning with {PyTorch
  Geometric}}. In \bibinfo{booktitle}{\emph{ICLR 2019 Workshop on
  Representation Learning on Graphs and Manifolds}} (New Orleans, USA).
\newblock
\urldef\tempurl%
\url{https://arxiv.org/abs/1903.02428}
\showURL{%
\tempurl}


\bibitem[Frasca et~al\mbox{.}(2020)]%
        {frasca2020sign}
\bibfield{author}{\bibinfo{person}{Fabrizio Frasca}, \bibinfo{person}{Emanuele
  Rossi}, \bibinfo{person}{Davide Eynard}, \bibinfo{person}{Ben Chamberlain},
  \bibinfo{person}{Michael Bronstein}, {and} \bibinfo{person}{Federico Monti}.}
  \bibinfo{year}{2020}\natexlab{}.
\newblock \showarticletitle{Sign: Scalable inception graph neural networks}.
\newblock \bibinfo{journal}{\emph{arXiv preprint arXiv:2004.11198}}
  (\bibinfo{year}{2020}).
\newblock


\bibitem[Hamilton et~al\mbox{.}(2017)]%
        {hamilton2017inductive}
\bibfield{author}{\bibinfo{person}{William~L Hamilton}, \bibinfo{person}{Rex
  Ying}, {and} \bibinfo{person}{Jure Leskovec}.}
  \bibinfo{year}{2017}\natexlab{}.
\newblock \showarticletitle{Inductive representation learning on large graphs}.
  In \bibinfo{booktitle}{\emph{NeurIPS}}. \bibinfo{pages}{1025--1035}.
\newblock


\bibitem[Hu et~al\mbox{.}(2021)]%
        {hu2021ogb}
\bibfield{author}{\bibinfo{person}{Weihua Hu}, \bibinfo{person}{Matthias Fey},
  \bibinfo{person}{Hongyu Ren}, \bibinfo{person}{Maho Nakata},
  \bibinfo{person}{Yuxiao Dong}, {and} \bibinfo{person}{Jure Leskovec}.}
  \bibinfo{year}{2021}\natexlab{}.
\newblock \showarticletitle{OGB-LSC: A Large-Scale Challenge for Machine
  Learning on Graphs}.
\newblock \bibinfo{journal}{\emph{arXiv preprint arXiv:2103.09430}}
  (\bibinfo{year}{2021}).
\newblock


\bibitem[Hu et~al\mbox{.}(2020)]%
        {hu2020heterogeneous}
\bibfield{author}{\bibinfo{person}{Ziniu Hu}, \bibinfo{person}{Yuxiao Dong},
  \bibinfo{person}{Kuansan Wang}, {and} \bibinfo{person}{Yizhou Sun}.}
  \bibinfo{year}{2020}\natexlab{}.
\newblock \showarticletitle{Heterogeneous graph transformer}. In
  \bibinfo{booktitle}{\emph{Proceedings of The Web Conference 2020}}.
  \bibinfo{pages}{2704--2710}.
\newblock


\bibitem[Huang et~al\mbox{.}(2020)]%
        {huang2020combining}
\bibfield{author}{\bibinfo{person}{Qian Huang}, \bibinfo{person}{Horace He},
  \bibinfo{person}{Abhay Singh}, \bibinfo{person}{Ser-Nam Lim}, {and}
  \bibinfo{person}{Austin~R Benson}.} \bibinfo{year}{2020}\natexlab{}.
\newblock \showarticletitle{Combining label propagation and simple models
  out-performs graph neural networks}.
\newblock \bibinfo{journal}{\emph{arXiv preprint arXiv:2010.13993}}
  (\bibinfo{year}{2020}).
\newblock


\bibitem[Huang et~al\mbox{.}(2018)]%
        {DBLP:conf/nips/Huang0RH18}
\bibfield{author}{\bibinfo{person}{Wen{-}bing Huang}, \bibinfo{person}{Tong
  Zhang}, \bibinfo{person}{Yu Rong}, {and} \bibinfo{person}{Junzhou Huang}.}
  \bibinfo{year}{2018}\natexlab{}.
\newblock \showarticletitle{Adaptive Sampling Towards Fast Graph Representation
  Learning}. In \bibinfo{booktitle}{\emph{NeurIPS}}.
  \bibinfo{pages}{4563--4572}.
\newblock


\bibitem[Jiang et~al\mbox{.}(2020)]%
        {jiang2020psgraph}
\bibfield{author}{\bibinfo{person}{Jiawei Jiang}, \bibinfo{person}{Pin Xiao},
  \bibinfo{person}{Lele Yu}, \bibinfo{person}{Xiaosen Li},
  \bibinfo{person}{Jiefeng Cheng}, \bibinfo{person}{Xupeng Miao},
  \bibinfo{person}{Zhipeng Zhang}, {and} \bibinfo{person}{Bin Cui}.}
  \bibinfo{year}{2020}\natexlab{}.
\newblock \showarticletitle{PSGraph: How Tencent trains extremely large-scale
  graphs with Spark?}. In \bibinfo{booktitle}{\emph{ICDE}}. IEEE,
  \bibinfo{pages}{1549--1557}.
\newblock


\bibitem[Jiang et~al\mbox{.}(2022)]%
        {jiang2022zoomer}
\bibfield{author}{\bibinfo{person}{Yuezihan Jiang}, \bibinfo{person}{Yu Cheng},
  \bibinfo{person}{Hanyu Zhao}, \bibinfo{person}{Wentao Zhang},
  \bibinfo{person}{Xupeng Miao}, \bibinfo{person}{Yu He},
  \bibinfo{person}{Liang Wang}, \bibinfo{person}{Zhi Yang}, {and}
  \bibinfo{person}{Bin Cui}.} \bibinfo{year}{2022}\natexlab{}.
\newblock \showarticletitle{ZOOMER: Boosting Retrieval on Web-scale Graphs by
  Regions of Interest}.
\newblock \bibinfo{journal}{\emph{arXiv preprint arXiv:2203.12596}}
  (\bibinfo{year}{2022}).
\newblock


\bibitem[Kipf and Welling(2016)]%
        {kipf2016semi}
\bibfield{author}{\bibinfo{person}{Thomas~N Kipf} {and} \bibinfo{person}{Max
  Welling}.} \bibinfo{year}{2016}\natexlab{}.
\newblock \showarticletitle{Semi-supervised classification with graph
  convolutional networks}.
\newblock \bibinfo{journal}{\emph{arXiv preprint arXiv:1609.02907}}
  (\bibinfo{year}{2016}).
\newblock


\bibitem[Klicpera et~al\mbox{.}(2019)]%
        {DBLP:conf/iclr/KlicperaBG19}
\bibfield{author}{\bibinfo{person}{Johannes Klicpera},
  \bibinfo{person}{Aleksandar Bojchevski}, {and} \bibinfo{person}{Stephan
  G{\"{u}}nnemann}.} \bibinfo{year}{2019}\natexlab{}.
\newblock \showarticletitle{Predict then Propagate: Graph Neural Networks meet
  Personalized PageRank}. In \bibinfo{booktitle}{\emph{ICLR}}.
\newblock


\bibitem[Li et~al\mbox{.}(2019)]%
        {li2019deepgcns}
\bibfield{author}{\bibinfo{person}{Guohao Li}, \bibinfo{person}{Matthias
  Muller}, \bibinfo{person}{Ali Thabet}, {and} \bibinfo{person}{Bernard
  Ghanem}.} \bibinfo{year}{2019}\natexlab{}.
\newblock \showarticletitle{Deepgcns: Can gcns go as deep as cnns?}. In
  \bibinfo{booktitle}{\emph{ICCV}}. \bibinfo{pages}{9267--9276}.
\newblock


\bibitem[Li et~al\mbox{.}(2018)]%
        {li2018deeper}
\bibfield{author}{\bibinfo{person}{Qimai Li}, \bibinfo{person}{Zhichao Han},
  {and} \bibinfo{person}{Xiao-Ming Wu}.} \bibinfo{year}{2018}\natexlab{}.
\newblock \showarticletitle{Deeper insights into graph convolutional networks
  for semi-supervised learning}. In \bibinfo{booktitle}{\emph{AAAI}},
  Vol.~\bibinfo{volume}{32}.
\newblock


\bibitem[Li et~al\mbox{.}(2021)]%
        {li2021openbox}
\bibfield{author}{\bibinfo{person}{Yang Li}, \bibinfo{person}{Yu Shen},
  \bibinfo{person}{Wentao Zhang}, \bibinfo{person}{Yuanwei Chen},
  \bibinfo{person}{Huaijun Jiang}, \bibinfo{person}{Mingchao Liu},
  \bibinfo{person}{Jiawei Jiang}, \bibinfo{person}{Jinyang Gao},
  \bibinfo{person}{Wentao Wu}, \bibinfo{person}{Zhi Yang}, {et~al\mbox{.}}}
  \bibinfo{year}{2021}\natexlab{}.
\newblock \showarticletitle{Openbox: A generalized black-box optimization
  service}. In \bibinfo{booktitle}{\emph{SIGKDD}}. \bibinfo{pages}{3209--3219}.
\newblock


\bibitem[Miao et~al\mbox{.}(2021a)]%
        {miao2021degnn}
\bibfield{author}{\bibinfo{person}{Xupeng Miao}, \bibinfo{person}{Nezihe~Merve
  G{\"u}rel}, \bibinfo{person}{Wentao Zhang}, {et~al\mbox{.}}}
  \bibinfo{year}{2021}\natexlab{a}.
\newblock \showarticletitle{Degnn: Improving graph neural networks with graph
  decomposition}. In \bibinfo{booktitle}{\emph{SIGKDD}}.
  \bibinfo{pages}{1223--1233}.
\newblock


\bibitem[Miao et~al\mbox{.}(2021b)]%
        {miao2021lasagne}
\bibfield{author}{\bibinfo{person}{Xupeng Miao}, \bibinfo{person}{Wentao
  Zhang}, \bibinfo{person}{Yingxia Shao}, \bibinfo{person}{Bin Cui},
  \bibinfo{person}{Lei Chen}, \bibinfo{person}{Ce Zhang}, {and}
  \bibinfo{person}{Jiawei Jiang}.} \bibinfo{year}{2021}\natexlab{b}.
\newblock \showarticletitle{Lasagne: A multi-layer graph convolutional network
  framework via node-aware deep architecture}.
\newblock \bibinfo{journal}{\emph{TKDE}} (\bibinfo{year}{2021}).
\newblock


\bibitem[Schlichtkrull et~al\mbox{.}(2018)]%
        {schlichtkrull2018modeling}
\bibfield{author}{\bibinfo{person}{Michael Schlichtkrull},
  \bibinfo{person}{Thomas~N Kipf}, \bibinfo{person}{Peter Bloem},
  \bibinfo{person}{Rianne Van Den~Berg}, \bibinfo{person}{Ivan Titov}, {and}
  \bibinfo{person}{Max Welling}.} \bibinfo{year}{2018}\natexlab{}.
\newblock \showarticletitle{Modeling relational data with graph convolutional
  networks}. In \bibinfo{booktitle}{\emph{European semantic web conference}}.
  Springer, \bibinfo{pages}{593--607}.
\newblock


\bibitem[Sen et~al\mbox{.}(2008)]%
        {DBLP:journals/aim/SenNBGGE08}
\bibfield{author}{\bibinfo{person}{Prithviraj Sen}, \bibinfo{person}{Galileo
  Namata}, \bibinfo{person}{Mustafa Bilgic}, \bibinfo{person}{Lise Getoor},
  \bibinfo{person}{Brian Gallagher}, {and} \bibinfo{person}{Tina
  Eliassi{-}Rad}.} \bibinfo{year}{2008}\natexlab{}.
\newblock \showarticletitle{Collective Classification in Network Data}.
\newblock \bibinfo{journal}{\emph{{AI} Mag.}} \bibinfo{volume}{29},
  \bibinfo{number}{3} (\bibinfo{year}{2008}), \bibinfo{pages}{93--106}.
\newblock


\bibitem[Shchur et~al\mbox{.}(2018)]%
        {shchur2018pitfalls}
\bibfield{author}{\bibinfo{person}{Oleksandr Shchur},
  \bibinfo{person}{Maximilian Mumme}, \bibinfo{person}{Aleksandar Bojchevski},
  {and} \bibinfo{person}{Stephan G{\"u}nnemann}.}
  \bibinfo{year}{2018}\natexlab{}.
\newblock \showarticletitle{Pitfalls of graph neural network evaluation}.
\newblock \bibinfo{journal}{\emph{arXiv preprint arXiv:1811.05868}}
  (\bibinfo{year}{2018}).
\newblock


\bibitem[Shi et~al\mbox{.}(2020)]%
        {shi2020masked}
\bibfield{author}{\bibinfo{person}{Yunsheng Shi}, \bibinfo{person}{Zhengjie
  Huang}, \bibinfo{person}{Wenjin Wang}, \bibinfo{person}{Hui Zhong},
  \bibinfo{person}{Shikun Feng}, {and} \bibinfo{person}{Yu Sun}.}
  \bibinfo{year}{2020}\natexlab{}.
\newblock \showarticletitle{Masked label prediction: Unified message passing
  model for semi-supervised classification}.
\newblock \bibinfo{journal}{\emph{arXiv preprint arXiv:2009.03509}}
  (\bibinfo{year}{2020}).
\newblock


\bibitem[Spinelli et~al\mbox{.}(2020)]%
        {spinelli2020adaptive}
\bibfield{author}{\bibinfo{person}{Indro Spinelli}, \bibinfo{person}{Simone
  Scardapane}, {and} \bibinfo{person}{Aurelio Uncini}.}
  \bibinfo{year}{2020}\natexlab{}.
\newblock \showarticletitle{Adaptive propagation graph convolutional network}.
\newblock \bibinfo{journal}{\emph{IEEE Transactions on Neural Networks and
  Learning Systems}} (\bibinfo{year}{2020}).
\newblock


\bibitem[Sun and Wu(2021)]%
        {sun2021scalable}
\bibfield{author}{\bibinfo{person}{Chuxiong Sun} {and} \bibinfo{person}{Guoshi
  Wu}.} \bibinfo{year}{2021}\natexlab{}.
\newblock \showarticletitle{Scalable and Adaptive Graph Neural Networks with
  Self-Label-Enhanced training}.
\newblock \bibinfo{journal}{\emph{arXiv preprint arXiv:2104.09376}}
  (\bibinfo{year}{2021}).
\newblock


\bibitem[Trouillon et~al\mbox{.}(2017)]%
        {trouillon2017knowledge}
\bibfield{author}{\bibinfo{person}{Th{\'e}o Trouillon},
  \bibinfo{person}{Christopher~R Dance}, \bibinfo{person}{Johannes Welbl},
  \bibinfo{person}{Sebastian Riedel}, \bibinfo{person}{{\'E}ric Gaussier},
  {and} \bibinfo{person}{Guillaume Bouchard}.} \bibinfo{year}{2017}\natexlab{}.
\newblock \showarticletitle{Knowledge graph completion via complex tensor
  factorization}.
\newblock \bibinfo{journal}{\emph{arXiv preprint arXiv:1702.06879}}
  (\bibinfo{year}{2017}).
\newblock


\bibitem[Veli{\v{c}}kovi{\'c} et~al\mbox{.}(2017)]%
        {velivckovic2017graph}
\bibfield{author}{\bibinfo{person}{Petar Veli{\v{c}}kovi{\'c}},
  \bibinfo{person}{Guillem Cucurull}, \bibinfo{person}{Arantxa Casanova},
  \bibinfo{person}{Adriana Romero}, \bibinfo{person}{Pietro Lio}, {and}
  \bibinfo{person}{Yoshua Bengio}.} \bibinfo{year}{2017}\natexlab{}.
\newblock \showarticletitle{Graph attention networks}.
\newblock \bibinfo{journal}{\emph{arXiv preprint arXiv:1710.10903}}
  (\bibinfo{year}{2017}).
\newblock


\bibitem[Wang et~al\mbox{.}(2021b)]%
        {wang2021flexgraph}
\bibfield{author}{\bibinfo{person}{Lei Wang}, \bibinfo{person}{Qiang Yin},
  \bibinfo{person}{Chao Tian}, \bibinfo{person}{Jianbang Yang},
  \bibinfo{person}{Rong Chen}, \bibinfo{person}{Wenyuan Yu},
  \bibinfo{person}{Zihang Yao}, {and} \bibinfo{person}{Jingren Zhou}.}
  \bibinfo{year}{2021}\natexlab{b}.
\newblock \showarticletitle{FlexGraph: a flexible and efficient distributed
  framework for GNN training}. In \bibinfo{booktitle}{\emph{Proceedings of the
  Sixteenth European Conference on Computer Systems}}. \bibinfo{pages}{67--82}.
\newblock


\bibitem[Wang et~al\mbox{.}(2021a)]%
        {wang2021bag}
\bibfield{author}{\bibinfo{person}{Yangkun Wang}, \bibinfo{person}{Jiarui Jin},
  \bibinfo{person}{Weinan Zhang}, \bibinfo{person}{Yong Yu},
  \bibinfo{person}{Zheng Zhang}, {and} \bibinfo{person}{David Wipf}.}
  \bibinfo{year}{2021}\natexlab{a}.
\newblock \showarticletitle{Bag of Tricks for Node Classification with Graph
  Neural Networks}.
\newblock \bibinfo{journal}{\emph{arXiv preprint arXiv:2103.13355}}
  (\bibinfo{year}{2021}).
\newblock


\bibitem[Wu et~al\mbox{.}(2019)]%
        {wu2019simplifying}
\bibfield{author}{\bibinfo{person}{Felix Wu}, \bibinfo{person}{Tianyi Zhang},
  \bibinfo{person}{Amauri Holanda~de Souza~Jr}, \bibinfo{person}{Christopher
  Fifty}, \bibinfo{person}{Tao Yu}, {and} \bibinfo{person}{Kilian~Q
  Weinberger}.} \bibinfo{year}{2019}\natexlab{}.
\newblock \showarticletitle{Simplifying graph convolutional networks}.
\newblock \bibinfo{journal}{\emph{arXiv preprint arXiv:1902.07153}}
  (\bibinfo{year}{2019}).
\newblock


\bibitem[Wu et~al\mbox{.}(2021)]%
        {wu2021r}
\bibfield{author}{\bibinfo{person}{Xinliang Wu}, \bibinfo{person}{Mengying
  Jiang}, {and} \bibinfo{person}{Guizhong Liu}.}
  \bibinfo{year}{2021}\natexlab{}.
\newblock \showarticletitle{R-GSN: The Relation-based Graph Similar Network for
  Heterogeneous Graph}.
\newblock \bibinfo{journal}{\emph{arXiv preprint arXiv:2103.07877}}
  (\bibinfo{year}{2021}).
\newblock


\bibitem[Xu et~al\mbox{.}(2018)]%
        {xu2018representation}
\bibfield{author}{\bibinfo{person}{Keyulu Xu}, \bibinfo{person}{Chengtao Li},
  \bibinfo{person}{Yonglong Tian}, \bibinfo{person}{Tomohiro Sonobe},
  \bibinfo{person}{Ken-ichi Kawarabayashi}, {and} \bibinfo{person}{Stefanie
  Jegelka}.} \bibinfo{year}{2018}\natexlab{}.
\newblock \showarticletitle{Representation learning on graphs with jumping
  knowledge networks}. In \bibinfo{booktitle}{\emph{ICML}}. PMLR,
  \bibinfo{pages}{5453--5462}.
\newblock


\bibitem[Yu et~al\mbox{.}(2020a)]%
        {yu2020scalable}
\bibfield{author}{\bibinfo{person}{Lingfan Yu}, \bibinfo{person}{Jiajun Shen},
  \bibinfo{person}{Jinyang Li}, {and} \bibinfo{person}{Adam Lerer}.}
  \bibinfo{year}{2020}\natexlab{a}.
\newblock \showarticletitle{Scalable graph neural networks for heterogeneous
  graphs}.
\newblock \bibinfo{journal}{\emph{arXiv preprint arXiv:2011.09679}}
  (\bibinfo{year}{2020}).
\newblock


\bibitem[Yu et~al\mbox{.}(2020b)]%
        {yu2020hybrid}
\bibfield{author}{\bibinfo{person}{Le Yu}, \bibinfo{person}{Leilei Sun},
  \bibinfo{person}{Bowen Du}, \bibinfo{person}{Chuanren Liu},
  \bibinfo{person}{Weifeng Lv}, {and} \bibinfo{person}{Hui Xiong}.}
  \bibinfo{year}{2020}\natexlab{b}.
\newblock \showarticletitle{Hybrid Micro/Macro Level Convolution for
  Heterogeneous Graph Learning}.
\newblock \bibinfo{journal}{\emph{arXiv preprint arXiv:2012.14722}}
  (\bibinfo{year}{2020}).
\newblock


\bibitem[Yu et~al\mbox{.}(2021)]%
        {yu2021heterogeneous}
\bibfield{author}{\bibinfo{person}{Le Yu}, \bibinfo{person}{Leilei Sun},
  \bibinfo{person}{Bowen Du}, \bibinfo{person}{Chuanren Liu},
  \bibinfo{person}{Weifeng Lv}, {and} \bibinfo{person}{Hui Xiong}.}
  \bibinfo{year}{2021}\natexlab{}.
\newblock \showarticletitle{Heterogeneous Graph Representation Learning with
  Relation Awareness}.
\newblock \bibinfo{journal}{\emph{arXiv preprint arXiv:2105.11122}}
  (\bibinfo{year}{2021}).
\newblock


\bibitem[Zaharia et~al\mbox{.}(2010)]%
        {zaharia2010spark}
\bibfield{author}{\bibinfo{person}{Matei Zaharia}, \bibinfo{person}{Mosharaf
  Chowdhury}, \bibinfo{person}{Michael~J Franklin}, \bibinfo{person}{Scott
  Shenker}, {and} \bibinfo{person}{Ion Stoica}.}
  \bibinfo{year}{2010}\natexlab{}.
\newblock \showarticletitle{Spark: Cluster computing with working sets}. In
  \bibinfo{booktitle}{\emph{2nd USENIX Workshop on Hot Topics in Cloud
  Computing (HotCloud 10)}}.
\newblock


\bibitem[Zeng et~al\mbox{.}(2020)]%
        {DBLP:conf/iclr/ZengZSKP20}
\bibfield{author}{\bibinfo{person}{Hanqing Zeng}, \bibinfo{person}{Hongkuan
  Zhou}, \bibinfo{person}{Ajitesh Srivastava}, \bibinfo{person}{Rajgopal
  Kannan}, {and} \bibinfo{person}{Viktor~K. Prasanna}.}
  \bibinfo{year}{2020}\natexlab{}.
\newblock \showarticletitle{GraphSAINT: Graph Sampling Based Inductive Learning
  Method}. In \bibinfo{booktitle}{\emph{ICLR}}.
\newblock


\bibitem[Zhang et~al\mbox{.}(2021a)]%
        {zhang2021rod}
\bibfield{author}{\bibinfo{person}{Wentao Zhang}, \bibinfo{person}{Yuezihan
  Jiang}, \bibinfo{person}{Yang Li}, \bibinfo{person}{Zeang Sheng},
  \bibinfo{person}{Yu Shen}, \bibinfo{person}{Xupeng Miao},
  \bibinfo{person}{Liang Wang}, \bibinfo{person}{Zhi Yang}, {and}
  \bibinfo{person}{Bin Cui}.} \bibinfo{year}{2021}\natexlab{a}.
\newblock \showarticletitle{ROD: reception-aware online distillation for sparse
  graphs}. In \bibinfo{booktitle}{\emph{SIGKDD}}. \bibinfo{pages}{2232--2242}.
\newblock


\bibitem[Zhang et~al\mbox{.}(2020)]%
        {zhang2020reliable}
\bibfield{author}{\bibinfo{person}{Wentao Zhang}, \bibinfo{person}{Xupeng
  Miao}, \bibinfo{person}{Yingxia Shao}, \bibinfo{person}{Jiawei Jiang},
  \bibinfo{person}{Lei Chen}, \bibinfo{person}{Olivier Ruas}, {and}
  \bibinfo{person}{Bin Cui}.} \bibinfo{year}{2020}\natexlab{}.
\newblock \showarticletitle{Reliable data distillation on graph convolutional
  network}. In \bibinfo{booktitle}{\emph{SIGMOD}}. \bibinfo{pages}{1399--1414}.
\newblock


\bibitem[Zhang et~al\mbox{.}(2022)]%
        {zhang2022pasca}
\bibfield{author}{\bibinfo{person}{Wentao Zhang}, \bibinfo{person}{Yu Shen},
  \bibinfo{person}{Zheyu Lin}, \bibinfo{person}{Yang Li},
  \bibinfo{person}{Xiaosen Li}, \bibinfo{person}{Wen Ouyang},
  \bibinfo{person}{Yangyu Tao}, \bibinfo{person}{Zhi Yang}, {and}
  \bibinfo{person}{Bin Cui}.} \bibinfo{year}{2022}\natexlab{}.
\newblock \showarticletitle{Pasca: A graph neural architecture search system
  under the scalable paradigm}. In \bibinfo{booktitle}{\emph{Proceedings of the
  ACM Web Conference 2022}}. \bibinfo{pages}{1817--1828}.
\newblock


\bibitem[Zhang et~al\mbox{.}(2021b)]%
        {zhang2021node}
\bibfield{author}{\bibinfo{person}{Wentao Zhang}, \bibinfo{person}{Mingyu
  Yang}, \bibinfo{person}{Zeang Sheng}, \bibinfo{person}{Yang Li},
  \bibinfo{person}{Wen Ouyang}, \bibinfo{person}{Yangyu Tao},
  \bibinfo{person}{Zhi Yang}, {and} \bibinfo{person}{Bin Cui}.}
  \bibinfo{year}{2021}\natexlab{b}.
\newblock \showarticletitle{Node Dependent Local Smoothing for Scalable Graph
  Learning}.
\newblock \bibinfo{journal}{\emph{NeurIPS}} (\bibinfo{year}{2021}).
\newblock


\bibitem[Zheng et~al\mbox{.}(2020)]%
        {distdgl_ai3_2020}
\bibfield{author}{\bibinfo{person}{Da Zheng}, \bibinfo{person}{Chao Ma},
  \bibinfo{person}{Minjie Wang}, \bibinfo{person}{Jinjing Zhou},
  \bibinfo{person}{Qidong Su}, \bibinfo{person}{Xiang Song},
  \bibinfo{person}{Quan Gan}, \bibinfo{person}{Zheng Zhang}, {and}
  \bibinfo{person}{George Karypis}.} \bibinfo{year}{2020}\natexlab{}.
\newblock \showarticletitle{DistDGL: Distributed Graph Neural Network Training
  for Billion-Scale Graphs}. In \bibinfo{booktitle}{\emph{10th {IEEE/ACM}
  Workshop on Irregular Applications: Architectures and Algorithms, {IA3} 2020,
  Atlanta, GA, USA, November 11, 2020}}. \bibinfo{publisher}{{IEEE}},
  \bibinfo{pages}{36--44}.
\newblock


\bibitem[Zhu and Koniusz(2021)]%
        {zhu2021simple}
\bibfield{author}{\bibinfo{person}{Hao Zhu} {and} \bibinfo{person}{Piotr
  Koniusz}.} \bibinfo{year}{2021}\natexlab{}.
\newblock \showarticletitle{Simple spectral graph convolution}. In
  \bibinfo{booktitle}{\emph{ICLR}}.
\newblock


\bibitem[Zhu et~al\mbox{.}(2019)]%
        {aligraph_vldb_2019}
\bibfield{author}{\bibinfo{person}{Rong Zhu}, \bibinfo{person}{Kun Zhao},
  \bibinfo{person}{Hongxia Yang}, \bibinfo{person}{Wei Lin},
  \bibinfo{person}{Chang Zhou}, \bibinfo{person}{Baole Ai},
  \bibinfo{person}{Yong Li}, {and} \bibinfo{person}{Jingren Zhou}.}
  \bibinfo{year}{2019}\natexlab{}.
\newblock \showarticletitle{AliGraph: A Comprehensive Graph Neural Network
  Platform}.
\newblock \bibinfo{journal}{\emph{Proc. VLDB Endow.}} \bibinfo{volume}{12},
  \bibinfo{number}{12} (\bibinfo{date}{Aug.} \bibinfo{year}{2019}),
  \bibinfo{pages}{2094–2105}.
\newblock
\showISSN{2150-8097}
\urldef\tempurl%
\url{https://doi.org/10.14778/3352063.3352127}
\showDOI{\tempurl}


\bibitem[Zhu and Ghahramani(2002)]%
        {zhu2002learnin}
\bibfield{author}{\bibinfo{person}{Xiaojin Zhu} {and} \bibinfo{person}{Zoubin
  Ghahramani}.} \bibinfo{year}{2002}\natexlab{}.
\newblock \showarticletitle{Learning from labeled and unlabeled data with label
  propagation}.
\newblock  (\bibinfo{year}{2002}).
\newblock


\end{thebibliography}

\clearpage

\appendix
\begin{figure}[tpb!]
    \vspace{-4mm}
	\centering
	\includegraphics[width=0.95\columnwidth]{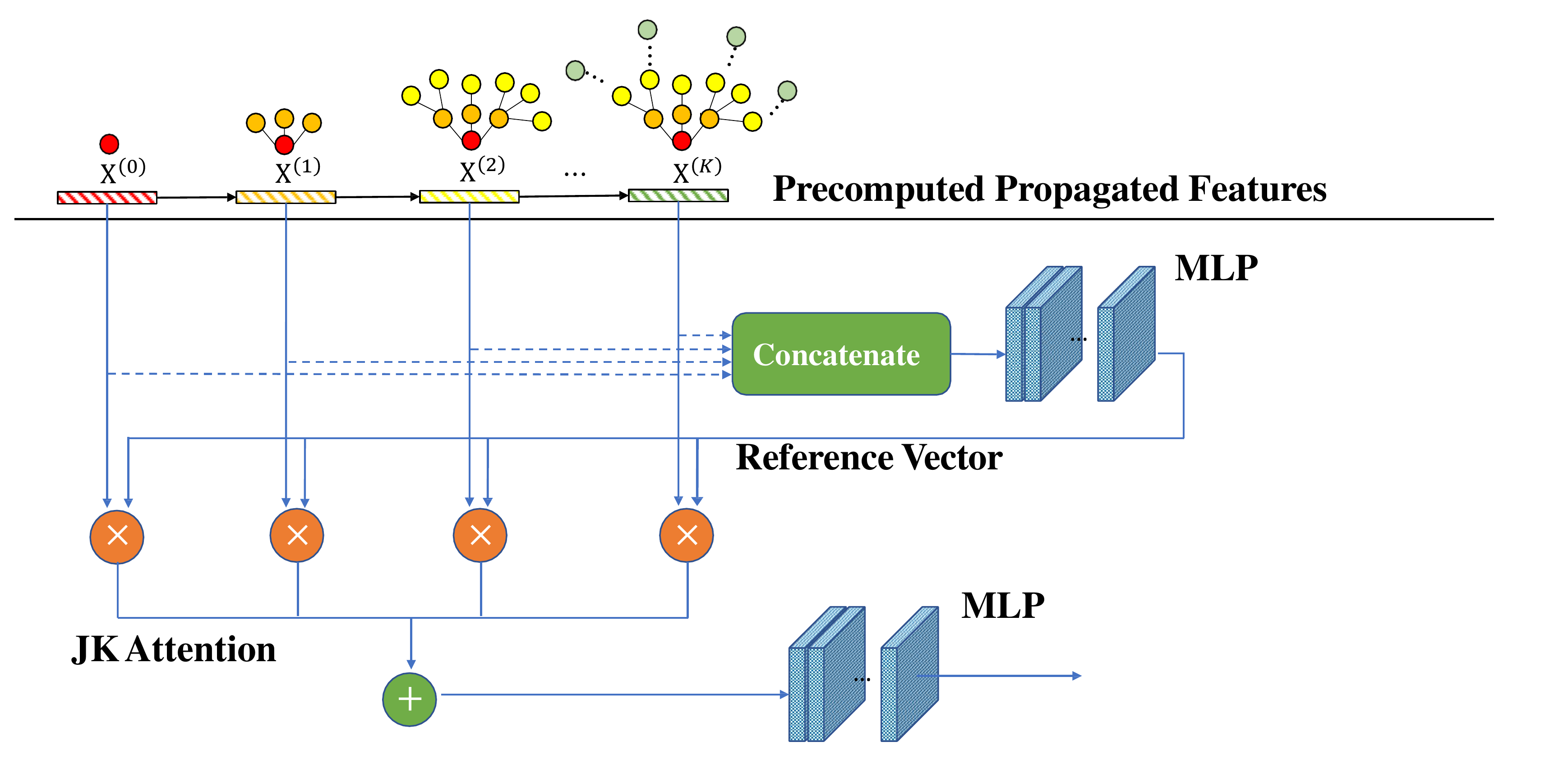}
	\caption{The architecture of GAMLP with JK Attention.}
	\label{fig:gmlp}
\end{figure}

\section{More details about GAMLP}
\subsection{An example of JK attention}
Fig.~\ref{fig:gmlp} provides a more zoomed-in look of JK attention, one of the two node-adaptive attention mechanisms we proposed.
The propagated features are concatenated and then fed into an MLP to map the concatenated feature to the hidden dimension of the model.
The mapped feature is then set as the reference vector of the following attention mechanism.
A linear layer is adopted to calculate the combination weight for propagated features at different propagation steps.
The propagated features are then multiplied with the corresponding combination weight, and the summed results are fed into another MLP to generate final predictions.

\subsection{Comparison between the label usage in UniMP and GAMLP}
(1) The label usage in UniMP is coupled with the training process, making it hard to scale to large graphs. While GAMLP decouples the label usage from the training process, the label propagation process can be executed as preprocessing. 

\noindent(2) The label propagation steps in UniMP are restricted to the same number of model layers. Moreover, UniMP will encounter the efficiency and scalability issues even on relatively small graphs if the number of model layers becomes large. 
In contrast, the label propagation steps in GAMLP can be quite large since the label propagation is performed as preprocessing. 

\noindent(3) Both propose approaches to fight against the label leakage issue. However, the random masking in UniMP has to be executed in each training epoch, while the last residual connection (composed of simple matrix addition) in GAMLP needs only to be executed once during preprocessing. Thus, UniMP consumes more resources than GAMLP to fight the label leakage issue.

\section{Additional experiments}
\label{more-exp}
\begin{figure}[tpb!]
    \centering
    \vspace{-4mm}
    \includegraphics[width=0.8\linewidth]{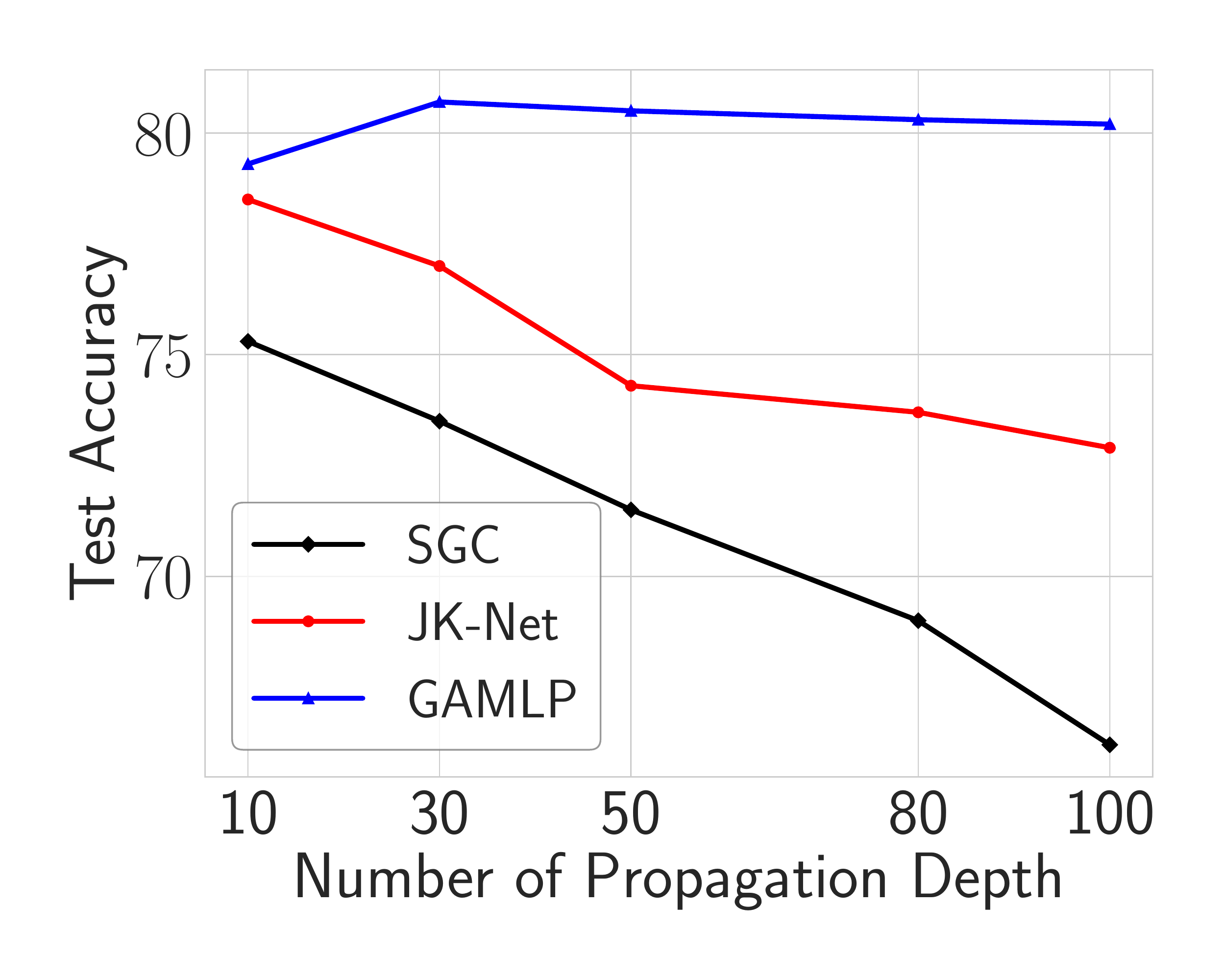}
    \vspace{-4mm}
    \caption{Test accuracy with different propagation depth.}
    \vspace{-4mm}
    \label{fig:deep_prop}
\end{figure}

\subsection{Deep propagation is possible }
\label{deep-pro}
Equipped with the learnable node-wise propagation scheme, our GAMLP can still maintain high predictive accuracy even when the propagation depth is over 50.
Here, we evaluate the predictive accuracy of our proposed GAMLP(JK) at propagation depth 10, 30, 50, 80, 100 on the PubMed dataset.
The performance of JK-Net and SGC are also reported as baselines.
The experimental results in Fig.~\ref{fig:deep_prop} show that even at propagation depth equals 100, the predictive accuracy of our GAMLP(JK) still exceeds 80.0\%, higher than the predictive accuracy of most baselines in Table~\ref{tab:transductive}.
At the same time, the predictive accuracy of SGC and JK-Net both drops rapidly when propagation depth increases from 10 to 100.

\begin{figure}[tpb!]
    \centering
    \includegraphics[width=0.8\linewidth]{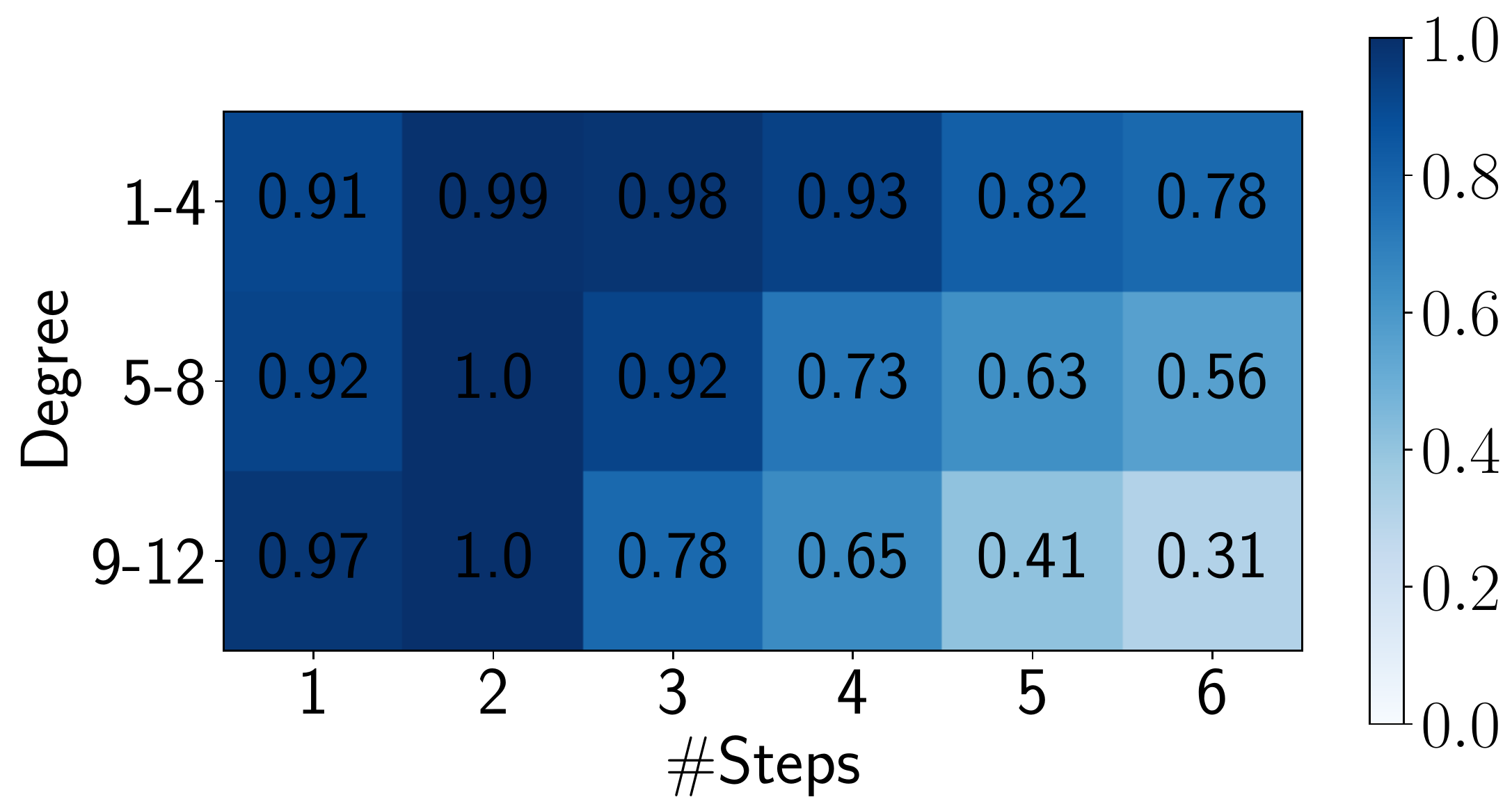}
    \vspace{-4mm}
    \caption{The average attention weights of propagated features of different steps on 60 randomly selected nodes from ogbn-products. }
    \vspace{-4mm}
    \label{interpretability}
\end{figure}

\subsection{Interpretability of the attention mechanism}
GAMLP can adaptively and effectively combine multi-scale propagated features for each node.
To demonstrate this, Fig.~\ref{interpretability} shows the average attention weights of propagated features of GAMLP(JK) according to the number of steps and degrees of input nodes, where the maximum step is 6. 
In this experiment, we randomly select 20 nodes for each degree range (1-4, 5-8, 9-12) and plot the relative weight based on the maximum value. 
We get two observations from the heat map: 1) The 1-step and 2-step propagated features are always of great importance, which shows that GAMLP captures the local information as those widely 2-layer methods do; 
2) The weights of propagated features with larger steps drop faster as the degree grows, indicating that our attention mechanism could prevent high-degree nodes from including excessive irrelevant nodes, leading to over-smoothing. 
From the two observations, we conclude that GAMLP can identify the different RF demands of nodes and explicitly weight each propagated feature.

\begin{table*}[t]
\centering
\caption{Overview of the 14 Datasets} \label{tab:data}
\resizebox{0.85\linewidth}{!}{
\begin{tabular}{ccccccccc}
\toprule
\textbf{Dataset}&\textbf{\#Nodes}& \textbf{\#Features}&\textbf{\#Edges}&\textbf{\#Classes}&\textbf{\#Train/Val/Test}&\textbf{Task type}&\textbf{Description}\\
\midrule
Cora& 2,708 & 1,433 &5,429&7& 140/500/1000 & Transductive&citation network\\
Citeseer& 3,327 & 3,703&4,732&6& 120/500/1000 & Transductive&citation network\\
Pubmed& 19,717 & 500 &44,338&3& 60/500/1000 & Transductive&citation network\\
\midrule
Amazon Computer& 13,381  & 767& 245,778 & 10 &200/300/12881&Transductive&co-purchase graph\\
Amazon Photo &7,487  & 745& 119,043 & 8 & 160/240/7,087&Transductive&co-purchase graph\\
Coauthor CS& 18,333  & 6,805 & 81,894 & 15& 300/450/17,583&Transductive&co-authorship graph \\
Coauthor Physics& 34,493 & 8,415 & 247,962 & 5& 100/150/34,243&Transductive&co-authorship graph \\
\midrule
ogbn-products & 2,449,029 & 100 & 61,859,140 & 47 & 196k/49k/2204k & Transductive & co-purchase graph \\
ogbn-papers100M & 111,059,956 & 128 & 1,615,685,872 & 172 & 1207k/125k/214k & Transductive & citation network \\
ogbn-mag & 1,939,743 & 128 & 21,111,007 & 349 & 626k/66k/37k & Transductive & citation network \\
\midrule
Tencent Video & 1,000,000 & 64 & 1,434,382 & 253 & 5k/10k/30k&Transductive&user-video graph\\
\midrule
PPI & 56,944 &50  &818,716  &  121&  45k / 6k / 6k  & Inductive & protein interactions network \\
Flickr& 89,250 & 500 & 899,756 & 7 &  44k/22k/22k & Inductive &image network\\
Reddit& 232,965 & 602 & 11,606,919 & 41 &  155k/23k/54k & Inductive & social network \\
\bottomrule
\end{tabular}}
\end{table*}

\subsection{Choices for $\alpha_l$ in Last Residual Connection}
Our first choice for the $\alpha_l$ in the last residual connection module is $\alpha_l=\frac{L-l}{L}$. However, we find that GAMLP still encounters the over-fitting issue on some datasets. Thus, we instead choose $\alpha_l=\cos(\frac{\pi l}{2L})$ to give more penalties to labels at large propagation steps.
We provide the performance comparison on the ogbn-products dataset in Table~\ref{table.ablation_last_residual}. Three weighting schemes for the last residual connection module are tested: "Cosine function" stands for $\alpha_l=\cos(\frac{\pi l}{2L})$, the one in GAMLP; "Linear-decreasing weight" stands for $\alpha_l=\frac{L-l}{L}$; and "Fixed weight" stands for $\alpha_l=0.7$. Table~\ref{table.ablation_last_residual} shows that the weighting scheme GAMLP adopts, $\alpha_l=\cos(\frac{\pi l}{2L})$, outperforms the other two options.

\subsection{Efficiency Comparison on ogbn-products}
We compare the efficiency of GAMLP with sampling-based GraphSAINT and Cluster-GCN, graph-wise-propagation-based SGC, and layer-wise-propagation-based SIGN on the ogbn-products dataset.
The results in Table~\ref{tab:efficiency2} illustrates that (1) sampling-based methods (e.g., GraphSAINT) consume much more time than graph/layer-wise-propagation based methods (e.g., SGC, SIGN) due to the high computation cost introduced by the sampling process;
(2) the two variants of GAMLP achieve the best predictive accuracy while requiring comparable training time with SGC.

\begin{table}[tpb!]
\caption{Ablation study of choices for $\alpha_l$ on ogbn-products.}
%\vspace{-2mm}
\centering
{
\noindent
\renewcommand{\multirowsetup}{\centering}
\resizebox{0.8\linewidth}{!}{
\begin{tabular}{c|c}
\toprule
\textbf{Choices} & \textbf{Test Accuracy} \\
\midrule
Fixed weight & 82.56$\pm$0.43 \\
Linear-decreasing weight & 82.72$\pm$0.93 \\
Cosine function & 83.59$\pm$0.05 \\
\bottomrule
\end{tabular}}}
\label{table.ablation_last_residual}
%\vspace{-3mm}
\end{table}

\section{Detailed experiment setup}
\label{app:settings}

\subsection{Experiment Environment}
\label{app:dataset}
We provide detailed information about the datasets we adopted during the experiment in Table~\ref{tab:data}.
To alleviate the influence of randomness, we repeat each method ten times and report the mean performance and the standard deviations.
For the largest ogbn-papers100M dataset, we run each method five times instead.
The experiments are conducted on a machine with Intel(R) Xeon(R) Platinum 8255C CPU@2.50GHz, and a single Tesla V100 GPU with 32GB GPU memory. 
The operating system of the machine is Ubuntu 16.04. 
As for software versions, we use Python 3.6, Pytorch 1.7.1, and CUDA 10.1.
The hyper-parameters in each baseline are set according to the original paper if available. Please refer to Appendix~\ref{hyperparameters} for the detailed hyperparameter settings for our GAMLP. Besides, the source code of the PyTorch implementation of \sys can be found in Github (\blue{\url{https://github.com/PKU-DAIR/GAMLP}}). 

\subsection{Detailed Hyperparameters}
\label{hyperparameters}
We provide the detailed hyperparameter setting on GAMLP in Table~\ref{table.parameter1},~\ref{table.parameter2} and~\ref{table.parameter3} to help reproduce the results.
The hyperparameters are tuned with the toolkit OpenBox~\citep{li2021openbox} or follow the settings in their original paper.
To reproduce the experimental results of GAMLP, just follow the same hyperparameter setting yet only run the python codes.

\begin{table}[tpb!]
% \vspace{-5mm}
\centering
{
\noindent
\caption{Efficiency comparison on the ogbn-products dataset}.
\label{tab:efficiency2}
\resizebox{0.95\linewidth}{!}{
\begin{tabular}{c|cccccc}
\toprule
\textbf{Methods} & SGC & SIGN & GAMLP(JK) & GAMLP(R) & GraphSAINT & Cluster-GCN \\ \midrule
Training time & 1.0 & 4.0  & 8.0 & 9.3  & 364  & 503 \\ 
Test accuracy  & 75.87 & 80.52  & 83.54  & \textbf{83.59}  & 79.08  & 78.97  \\ \bottomrule
\end{tabular}}}
\label{table.effi-ogbn}
\vspace{1mm}
\end{table}

\begin{table}[tpb!]
\caption{Detailed hyperparameter setting on OGB datasets.}
%\vspace{-2mm}
\centering
{
\noindent
\renewcommand{\multirowsetup}{\centering}
\resizebox{0.95\linewidth}{!}{
\begin{tabular}{|c|c|c|c|c|c|}
\hline
\textbf{Datasets} & \textbf{\makecell{attention \\type}} & \textbf{\makecell{hidden \\ size}} & \textbf{\makecell{num layer \\ in JK}} & \textbf{num layer}  & \textbf{activation} \\ \hline
ogbn-products   & Recursive & 512  & / & 4 & leaky relu, a=0.2  \\ \hline
ogbn-papers100M & JK & 1280  & 4 & 6 & sigmoid \\ \hline
ogbn-mag        & JK  & 512  & 4 & 4 & leaky relu, a=0.2\\ \hline
\end{tabular}}}
\label{table.parameter1}
%\vspace{-3mm}
\end{table}

\begin{table}[tpb!]
\caption{Detailed hyperparameter setting on OGB datasets.}
%\vspace{-2mm}
\centering
{
\noindent
\renewcommand{\multirowsetup}{\centering}
\resizebox{0.95\linewidth}{!}{
\begin{tabular}{|c|c|c|c|c|c|}
\hline
\textbf{Datasets}   & \textbf{hops} & \textbf{\makecell{hops for\\ label}} & \textbf{\makecell{input \\dropout}} & \textbf{\makecell{attention \\dropout}} & \textbf{dropout}           \\ \hline
ogbn-products   & 5 & 10 & 0.2 & 0.5 & 0.5 \\ \hline
ogbn-papers100M & 12 & 10 & 0 & 0.5 & 0.5\\ \hline
ogbn-mag        & 5 & 3 & 0.1  & 0 & 0.5 \\ \hline
\end{tabular}}}
\label{table.parameter2}
%\vspace{-3mm}
\end{table}

\begin{table}[tpb!]
\caption{Detailed hyperparameter setting on OGB datasets.}
%\vspace{-2mm}
\centering
{
\noindent
\renewcommand{\multirowsetup}{\centering}
\resizebox{0.9\linewidth}{!}{
\begin{tabular}{|c|c|c|c|c|c|c|c|}
\hline
\textbf{Datasets}   & \textbf{beta} & \textbf{patience} & \textbf{lr} & \textbf{batch size} & \textbf{epochs}         \\ \hline
ogb-products    & 1 & 300  & 0.001 & 50000 & 400 \\ \hline
ogb-papers100M & 1 & 60 & 0.0001 & 5000 & 400 \\ \hline
ogb-mag        & 1 & 100  & 0.001  & 10000 & 400 \\ \hline
\end{tabular}}}
\label{table.parameter3}
%\vspace{-3mm}
\end{table}

\end{document}